\documentclass[journal]{IEEEtranTIE}
\usepackage{graphicx}
\usepackage{cite}
\usepackage{picinpar}
\usepackage{amsmath}
\usepackage{url}
\usepackage{flushend}
\usepackage[latin1]{inputenc}
\usepackage{colortbl}
\usepackage{soul}
\usepackage{multirow}
\usepackage{pifont}
\usepackage{color}
\usepackage{alltt}
\usepackage[hidelinks]{hyperref}
\usepackage{enumerate}
\usepackage{siunitx}
\usepackage{breakurl}
\usepackage{epstopdf}
\usepackage{pbox}
\usepackage{amsmath}
\usepackage{amssymb}
\usepackage{booktabs}
\usepackage{multirow}
\usepackage{multicol}
\usepackage{subfig}

\begin{document}
\clearpage
{
Please cite this paper as X. Sun, H. Xv, J. Dong, H. Zhou, C. Chen, and Q. Li, "Few-shot Learning for Domain-specific Fine-grained Image Classification," IEEE Transactions on Industrial Electronics, pp. 1-1, 2020.
}
\clearpage
	
\title{Few-shot Learning for Domain-specific Fine-grained Image Classification}

\author{
	\vskip 1em
	{
	Xin~Sun, \emph{Member, IEEE},
	Hongwei~Xv,
	Junyu~Dong, \emph{Member, IEEE},\\
	Huiyu~Zhou, \emph{Member, IEEE},
	Changrui~Chen,
	and Qiong~Li
	}

	\thanks{
		
		{
		Manuscript received Month xx, 2xxx; revised Month xx, xxxx; accepted Month x, xxxx.
		This work was supported in part by the National Natural Science Foundation of China (No. U1706218, 61971388, L1824025), Key Research and Development Program of Shandong Province (No. GG201703140154), and Major Program of Natural Science Foundation of Shandong Province (No. ZR2018ZB0852). H. Zhou was supported by Royal Society-Newton Advanced Fellowship under Grant NA160342, and European Union's Horizon 2020 research and innovation program under the Marie-Sklodowska-Curie grant agreement No 720325.		
	
		X. Sun, H. Xv, J. Dong, C. Chen, Q. Li are with the Department of Computer Science and Technology, Ocean University of China, Qingdao, PR.China. (e-mail: sunxin1984@ieee.org, dongjunyu@ouc.edu.cn).
		
		H. Zhou is with the School of Informatics, University of Leicester, UK. (e-mail: hz143@leicester.ac.uk)
		}
	}
}

\maketitle
	
\begin{abstract}
Learning to recognize novel visual categories from a few examples is a challenging task for machines in real-world industrial applications. In contrast, humans have the ability to discriminate even similar objects with little supervision. This paper attempts to address the few-shot fine-grained image classification problem. We propose a feature fusion model to explore discriminative features by focusing on key regions. The model utilizes the focus-area location mechanism to discover the perceptually similar regions among objects. High-order integration is employed to capture the interaction information among intra-parts. We also design a Center Neighbor Loss to form robust embedding space distributions. Furthermore, we build a typical fine-grained and few-shot learning dataset {\it{\textbf{mini}}}{\textbf{PPlankton}} from the real-world application in the area of marine ecological environments. Extensive experiments are carried out to validate the performance of our method. The results demonstrate that our model achieves competitive performance compared with state-of-the-art models. Our work is a valuable complement to the model domain-specific industrial applications.
\end{abstract}

\begin{IEEEkeywords}
Computer vision; Few-shot learning; Representation learning
\end{IEEEkeywords}

\markboth{IEEE TRANSACTIONS ON INDUSTRIAL ELECTRONICS}%
{}

\definecolor{limegreen}{rgb}{0.2, 0.8, 0.2}
\definecolor{forestgreen}{rgb}{0.13, 0.55, 0.13}
\definecolor{greenhtml}{rgb}{0.0, 0.5, 0.0}

\section{Introduction}

\IEEEPARstart{I}{n recent years,} we have witnessed significant progress in computer vision \cite{He2017, Zhuang2012Multichannel}. Thanks to large-scale of labeled training data, e.g., ImageNet, deep convolutional neural networks (ConvNets) are able to successfully learn robust feature representations and achieve excellent performance in recognition tasks. Although it has high accuracy in various labeled datasets, the generalization ability of the ConvNet model is still weak. In particular, the ConvNet model is difficult to quickly identify a novel category using only one or a few labeled samples. However, humans are able to recognize new objects easily with very little supervision \cite{Lake2015}. For example, kids have no problem to generalize the concept of ``panda'' from only one picture. Furthermore, experts will be faster to understand novel concepts with prior professional knowledge. This work focuses on the task that recognizing novel visual categories after seeing just a few labeled examples. Research on this subject is often termed \emph {few-shot learning}. 


In contrast to the common image classification problem in daily life, most of the real-world scenarios face few-shot problems. For example, marine biologist pays great attention to the phytoplankton recognition problem which is a typical fine-grained and few-shot learning issue. The change of their abundance, e.g. eutrophication, is a significant indicator of the oceanic ecosystem's health. It is therefore very important to automatically identify phytoplankton in a certain area of the ocean. However, collections of phytoplankton images are very difficult. It is commonly accomplished by professional instruments such as electron microscope. Only a few samples of valuable categories can be discovered in one expensive sampling task. Therefore, the fine-grained and few-shot model is critical for domain-specific issues and has become one of the important topics in computer vision.

\begin{figure}
	\begin{center}
		\includegraphics[width=1\linewidth]{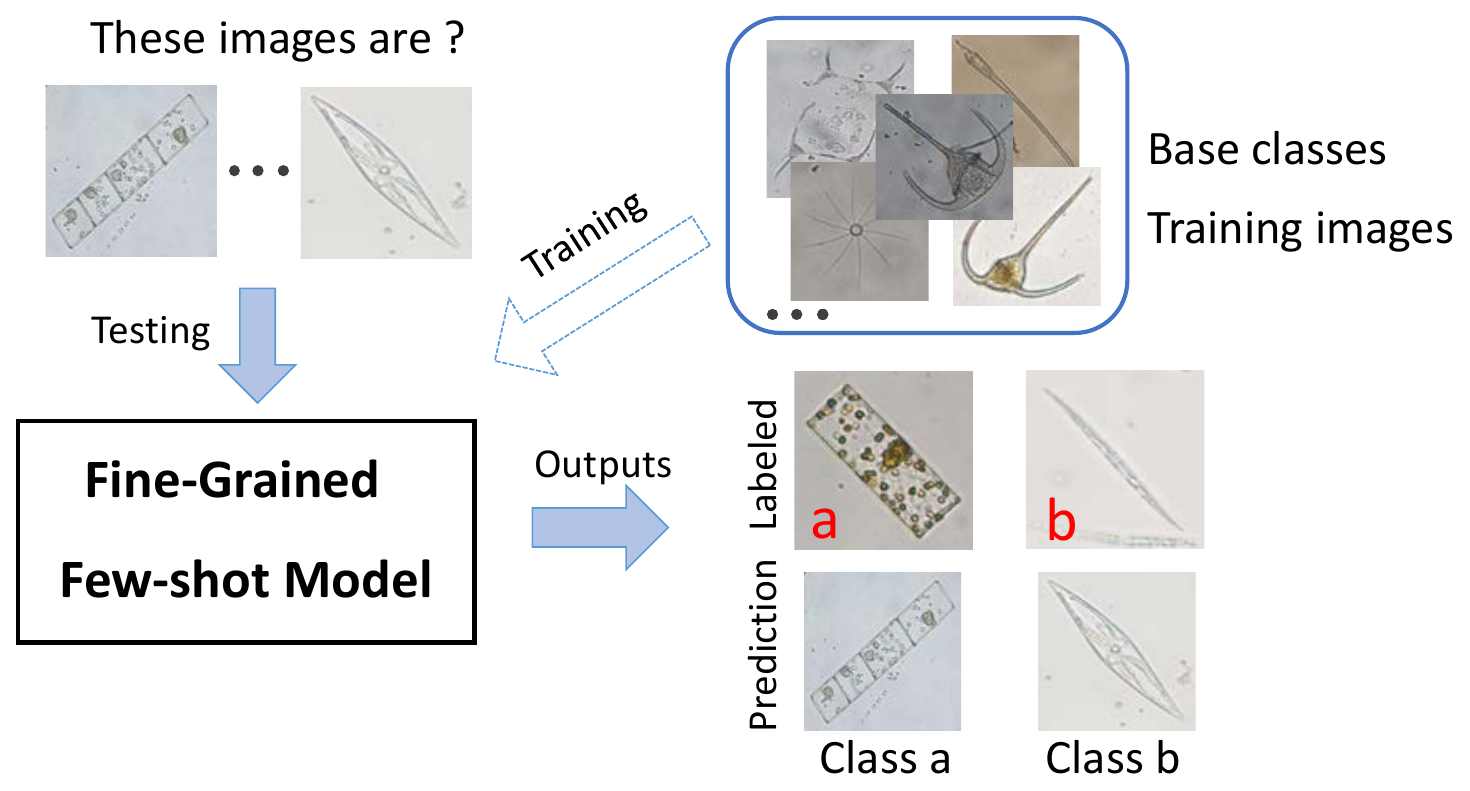}
	\end{center}
	\caption{A brief illustration of fine-grained few-shot recognition.}
	\label{start}
\end{figure}
Most of the few-shot learning methods fall under the umbrella of metric-learning. The metric-learning approaches try to solve these problems by placing new classes in a metric space (e.g., Euclidean or cosine distances) that can easily separate classes. For instance, Matching Networks \cite{Vinyals2016} can be interpreted as a nearest-neighbor classifier which can be trained end-to-end over the cosine distance. Notably, the training procedure has to be chosen carefully so as to match inference at the test stage. Each episode is designed to mimic the few-shot tasks by subsampling classes as well as data points (e.g., every episode sampling 5 classes and each class has 5 labeled samples). Prototypical Networks \cite{Snell2017} handles the few-shot tasks by calculating the Euclidean distance between the embedding points of query set and prototype representation of support set. Meanwhile, the pre-defined metric is no longer used in Relation Networks \cite{Sung2017}. It uses concatenated feature maps from the query and support images to distinguish similar and dissimilar samples.

It is very important to explore the relationship between feature representation of template images and that of the query image. Thus, to succeed in few-shot metric tasks, we shall make sure two aspects. First, we shall have a well-trained feature extractor. The other is an effective classifier including good metrics.	However, the above-mentioned methods are not conducive to ConvNets for extracting robust features and can sacrifice the accuracy of initial categories \cite{Gidaris2018}. Most of the few-shot methods pay attention to learning a deep distance metric to compare query images with the labeled images, while ignoring the importance of mining the better features from the existing few categories. That means it is critical to mining rich information from the labeled samples of few categories. Motivated by the above observation, we propose a Feature Fusion Model for obtaining more discriminative information from focus areas. We also design a loss function (Center Neighbor Loss) to help the whole architecture to learn better feature space distributions.

For the special fine-grained few-shot visual problem, we further build a microimage dataset of phytoplankton, i.e., {\it{\textbf{mini}}}{\textbf{PPlankton}. Unlike toy datasets for few-shot learning in literature, 
	 the {\it{\textbf{mini}}}{\textbf{PPlankton} dataset comes from the real-world tasks and can be used to evaluate fine-grained and few-shot methods. It illustrates a typical fine-grained and few-shot problem in marine biological science.
		
The main contributions of this paper are as follows:

\begin{enumerate}[1)]
	\item We propose a feature fusion model to explore the features by focusing on the key regions. It utilizes the focus-area location mechanism to discover the similarity regions between objects. Meanwhile, high-order integration is used to capture the intra-parts discriminative information.
	\item We design a Center Neighbor Loss function to form robust feature space distributions for generating discriminative features, to accomplish the fine-grained few-shot visual categorization task.
	\item We build a domain-specific fine-grained and few-shot dataset {\it{\textbf{mini}}}{\textbf{PPlankton}} for the real-world phytoplankton recognition problem. Experiments on the {\it{\textbf{mini}}}{\textbf{PPlankton}} show the superiority of the proposed model compared with other models.
\end{enumerate}
The rest of this paper is organized as follows. Section II summarizes the related works. Section III formally describes our model. Section IV presents the experimental results. Finally, we conclude in Section V.

\section{Related work}

Deep convolutional neural networks have made significant achievements for a wide range of visual tasks \cite{8663605, Qi2017Automatic, chen2020r}. Nevertheless, for fine-grained image categorization \cite{Zhang2015A}, it remains quite challenging to obtain the discriminative representations. In particular, it is a novel challenge to classify fine-grained images using only a few labeled sample images . 
The convolutional neural networks usually require thousands of labeled examples of each class to saturate performance. However, it is impractical to collect large amounts of annotated data, especially the domain-specific industrial applications that requires expert knowledge, such as oceanography \cite{Gorsky2010, Jakobsen2011}.  Recently, there is a resurgence of interest on few-shot learning \cite{Vinyals2016, Snell2017, Sung2017}. And a few research works are already pay attention to the fine-grained few-shot visual problem \cite{Huang2019, Huang2019a, Pahde2018, Das2019}.

Among the recent literature of few-shot learning, the metric learning and attention mechanism are most relevant proposed method. Metric learning has been successfully applied to face recognition \cite{Sun2014} and fine-grained image classification \cite{Zhang2015A}. The core idea is to learn an embedding function that the samples of the same category are closer than those of different classes. Once the embedding function is learned, the query images will be classified. Siamese network \cite{Chopra2005} consists of two identical sub-ConvNets that minimize the distances between paired data with the same labels while keeping the distances with different labels far apart.  Triplet loss \cite{Taigman2015} attempts to focus on relative distances rather than absolute pair-wise distances. It has been widely implemented in fine-grained tasks \cite{Wang2014}. However, the problem of triplet loss is dramatic data expansion when selecting triplets. Furthermore, center loss \cite{Wen2016a} can obtain highly discriminative features for robust face recognition. And it is unnecessary to design the sampling strategy carefully as contrastive loss and triplet loss do. The center loss has shown benefits in face identification. However, its performance is unknown for the fine-grained few-shot tasks. Then we further design a Center Neighbor Loss for achieving a robust embedding space.

It is critical to know which part of the images worth paying attention to. To acquire the attention feature representation, Li et al. \cite{Li} proposed a zoom network which utilized the candidate region to crop the original images. Wei et al. \cite{Wei2019} adopted the unsupervised object discovery and co-localization mechanism by deep descriptor transformation to discover the attention area. The attention mechanism is a possible way for learning robust representation. In this work, we introduce the focus-area location mechanism Grad-CAM \cite{Selvaraju2017} to find regions with discriminative features, which are critical for fine-grained classification.

Few-shot learning is critical in model industrial applications, such as novel species discovering. In this work, we take one typical real-world industrial problem to verify our method, i.e., phytoplankton classification. Marine phytoplankton is the foundation of the marine ecosystem \cite{Charlson1987}. It is an ecological concept that refers to tiny plants that float in the water. Plankton image classification \footnote{We no longer distinguish the image classification of phytoplankton and zooplankton separately.} is becoming critically important for marine observations and aquaculture.
\begin{figure*}[ht]
	\begin{center}
		\includegraphics[width=1\linewidth]{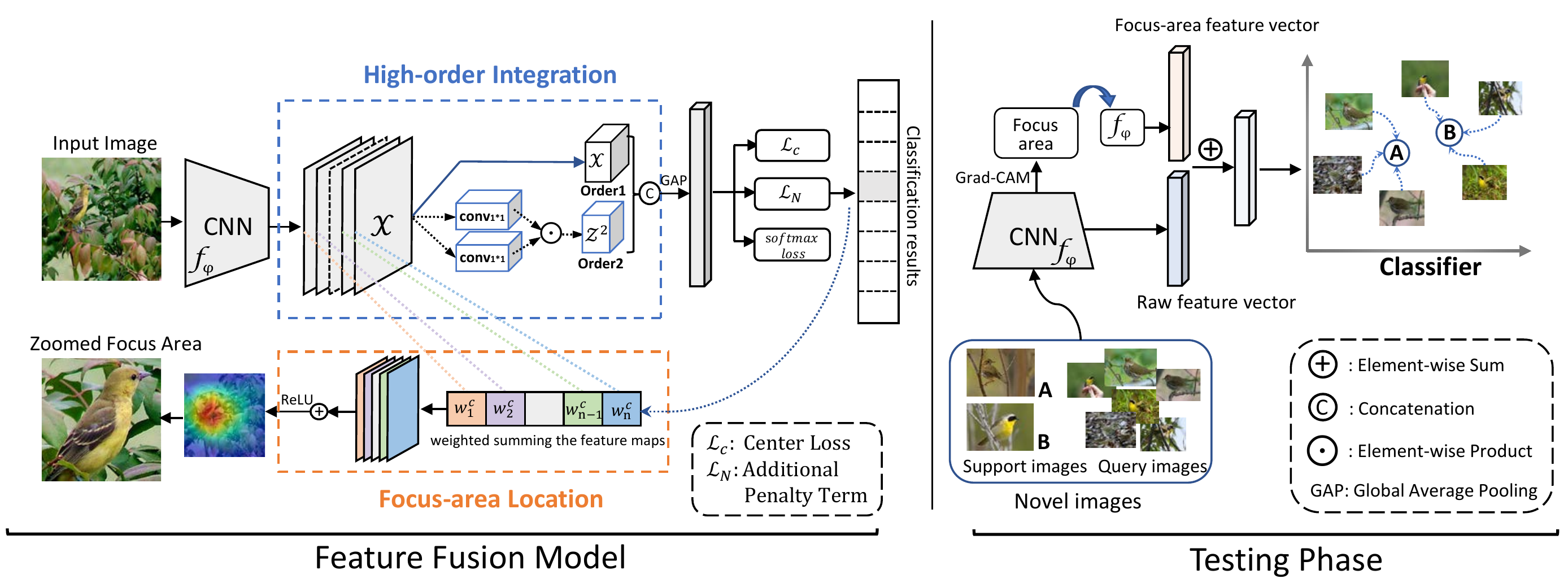}
	\end{center}
	\caption{The overview framework of our method. It consists of a {\it ConvNet-based feature extractor $f_{\varphi}$ }, a {\it feature fusion model} which is formed by focus-area location mechanism and high-order integration , and a {\it  cosine-similarity based classifier}. During the testing process, we classify unlabeled samples by comparing the cosine similarities of support set $\mathcal{S}$ and query set $\mathcal{Q}$.}
	
	\label{fig1}
\end{figure*}

The research of phytoplankton detection mainly relies on people to manually identify and count through the microscope. Current monitoring systems (e.g. ZooScan and FlowCAM \cite{Gorsky2010, Jakobsen2011}) yield large amounts of images every day. They are usually time-consuming, labor-intensive and needs strong professional knowledge. Schroder et al. \cite{Taylor2010} also notice the importance of classifying plankton only using a few labeled samples. They directly use weight Imprinting \cite{Brown} to enable a neural network to recognize small classes immediately without re-training.


\section{Methodology}
\subsection{Notation}
For few-shot classification,  there is a base train dataset $\mathcal{D}_{base}=\{(x_i, y_i)\}_{i=1}^{N}$ consisting of $N$ labeled images, where $y_i$ is the label of image $x_i$. Crucially, the model must distinguish a set of novel categories $\mathcal{Q} = \{(x_j, y_j)\}_{j=1}^{N_q}$ with a few training examples per category. These training examples are called support set, i.e., $\mathcal{S} = \{(x_ i^s,y_i^s)\}_{i=1}^{N_s} (N_s = K*C) $ which contains $K$ labeled examples for each of $C$ unique novel classes. $\mathcal{Q}$ acts as the unlabeled query set. Here $\mathcal{S} \bigcup \mathcal{Q}=\mathcal{D}_{novel}$ and  $\mathcal{D}_{base} \bigcap \mathcal{D}_{novel} = \emptyset$. This target few-shot task is named $C$-way $K$-shot.
\subsection{Model}
An overview of our method is illustrated in Fig.~\ref{fig1}, which mainly consists of three parts.

\subsubsection{ConvNet-based feature extractor}
A feature extractor $f_{\varphi}$, which parameterized by a ConvNet (e.g., ResNet \cite{He2016}), maps an input image $x\in \mathbb{R}^N$ to a $d$-dimensional feature vector $f_{\varphi}(x)\in\mathbb{R}^d$. As a classification model, $f_{\varphi}$ has a dot-product based classifier $C(.|W)$ (i.e., Last linear layer), where $W={\{w_i\in \mathbb{R}^d\}}_{i=1}^{K}$ is the set of weight vectors of the $K$ base classes. We can get the probability scores of the base training categories by calculating $C(f_{\varphi}(x) | W)$ and optimize the feature extractor by back-propagation.

\subsubsection{Feature fusion module}
For few-shot learning, it is pivotal to mine the largest support information from the support set $\mathcal{S}$. We propose a feature fusion model which utilizes the focus-area location and high-order integration to generate feature representation for the few-shot tasks. As shown in Fig.~\ref{fig1}, it consists of two components: (1) high-order integration, and (2) focus-area location.

{\bf High-order integration.} The recent progress of fine-grained classification demonstrates that the high-order representations with ConvNets can greatly improve its performance \cite{Cai, koniusz}. Intuitively, the key for fine-grained few-shot tasks is to represent the regions within same category that have a closer appearance and to exhibit discriminative areas between the different categories.

We assume that $\mathcal{X}\in\mathbb{R}^{K\times M\times N}$ is a 3$D$ feature map from the convolutional layers, where $x\in\mathcal{X}$ is a $K$-dimensional descriptor of one particular location region $p \in M\times N$. The linear predictor $\mathcal{W}$ on the high-order statistics of $\mathcal{X}$ could be formulated as follow.
\begin{equation}\label{kernel}
f(\mathcal{X}) = <\mathcal{W}, \sum\limits_{x\in\mathcal{X}} \phi(x)>
\end{equation}
where $\sum\limits_{x\in\mathcal{X}} \phi(x)$ denotes the high-order statistics characterized by a homogenous polynomial kernel \cite{Pham2013}. The  $\mathcal{W}$ can be approximated by rank-one decomposition. The tensor rank decomposition expresses a tensor as a minimum-length linear combination of rank-1 tensors. The outer product of vectors $ {{\bf{u}}_1} \in \mathbb{R}^{K_1},\ldots, {\bf{u}}_r \in {\mathbb{R}}^{K_r}$ is the $K_1 \times \ldots \times K_r $ rank-1 tensor that satisfies $({\bf{u}}_1 \otimes \ldots \otimes {\bf{u}}_r)_{k_1\ldots,k_r} = ({\bf{u}}_1)_{k_1} \ldots   ({\bf{u}}_r)_{k_r}$. The $\mathcal{W}$ can be rewritten as $\mathcal{W}=\sum_{d=1}^{D}a^d{{\bf{u}}_1^d}\otimes \ldots \otimes {{\bf{u}}_r^d} $, where $a^d$ is the weight for $d$-th rank-one tensor and $D$ is the rank of the tensor if D is minimal. Thus, Equation \ref{kernel} can be reformulated as follow.
\begin{equation} \label{kernelv2}
\begin{split}
f(\mathcal{X} )&=\sum\limits_{\mathrm{x} \in\mathcal{X}} \Bigg \{   	\left \langle \mathrm{ {\bf{w}}^1,\bf{x} } \right \rangle +\sum\limits_{r=2}^R \sum\limits_{d=1}^{D^r} a^{r,d} \prod_{s=1}^r  \left \langle { {\bf{u}}_s^{r,d}, \mathrm{x}}\right \rangle      \Bigg \},          \\
&=\left \langle \mathrm{ {\bf{w}}^1,\sum\limits_{x\in\mathcal{X}}  \bf{x} } \right \rangle + \sum\limits_{r=2}^R \left \langle { {\boldsymbol{a}}^r,     \sum\limits_{z^r\in\mathcal{Z}^r} {\boldsymbol{z}}^r } \right \rangle
\end{split}
\end{equation}  where the $\boldsymbol{z}^r = [ z^{r,1}, \dots, z^{r,D^r} ]^{\top}$ with $z^{r,d}= \prod_{s=1}^r{ \langle { {\bf{u}}_s^{r, d}, \mathrm{\bf{x}}  }  \rangle }$ characterizes the degree-$r$ variable interactions under a single rank-1 tensors, and $\boldsymbol{a}^r$ is the weight vector. The $\boldsymbol{z}^r$ can be calculated by performing $r$-th $1\times1$ convolutions with $D^r$ channel \cite{Wang1},
i.e., $\mathcal{Z}^r = \{\boldsymbol{z}^r\} = \prod_{i=1}^r  conv^{i}_{1\times1\times D^r}(\mathcal{X})$ .
In our feature fusion operation as shown in Fig.~\ref{fig1}, we integrate 2nd-order representations to capture more complex and high-order relationships among parts. After that, we perform global average pooling (GAP) \cite{Lin} to further aggregate features.

{\textbf{Focus-area location.} Existing studies show that learning from object regions could benefit object recognition at image-level \cite{Li}. Such focus-area in an image which benefit  few-shot learning. During the training procedure, $f_{\varphi}$ can generate focus-areas of images by Grad-CAM \cite{Selvaraju2017}, as formulated below.
	
	\begin{equation}\label{grad}
	L_{Grad-CAM}^c=ReLU(\sum\limits_k\alpha_k^cA^k)
	\end{equation}
	where $\alpha_k^c$ denotes the weight of the $k$-th feature map for category $c$. $\alpha_k^c$ can be calculated by the following formula.	
	\begin{equation}\label{gradv2}
	\alpha_k^c=\frac{1}{Z}\sum\limits_{i}\sum\limits_{j}\frac{\partial y^c}{\partial A_{ij}^k}
	\end{equation}
where $Z$ is the number of pixels in feature map, $y^c$ is the classification score corresponding to the category $c$, and $A_{ij}^k$ denotes the pixel value at the location of $(i,j)$ of the $k$-th feature map.
	
Grad-CAM has the ability to locate the focus areas that belong to the corresponding category. As shown in Fig.~\ref{fig1}, the dot line is a diagram of Grad-CAM, which represents the focus-area is obtained by weighted summing the feature maps. In this work, we utilize Grad-CAM to generate base categories' focus regions $\mathcal{H}_{base} = \{(x_i^h, y_i)\}_{i=1}^{N}$. However, the ConvNet extractor can not give a correct response of $c$ in formula (\ref{gradv2}), when a novel category appears. To our delight, we find that the model has accumulated lots of meta-knowledge in the domain field (e.g., Ornithology) during the training process of $\mathcal{D}_{base} = \{(x_i, y_i)\}_{i=1}^{N}$. The concepts of novel categories can be made up of various meta-knowledge, which are already embedded in the neural networks. For example, if someone has never seen the tiger, she/he might think it has many close parallels to a cat (learned before). The reason is that the attention locations of human on the new category tiger and the known category cat are similar to each other. Although we don't know the ground truth of the novel samples for the fine-grained few-shot tasks, the unseen class always has similar regions to the $\mathcal{D}_{base}$, such as bird's mouths and wings. And the base classifier will classify the new sample into the most similar class in $\mathcal{D}_{base}$. Therefore, it is possible to utilize Grad-CAM tp generate good focus-area location $\mathcal{H}_{novel}$ on the unseen categories for enhancing feature representation.
	
Telling the neural network the regions of rich discriminative information will form a more robust representation. This step is similar to the data augmentation of input space. However we only mine the available information on the input data itself without using the extra data augmentation.

\subsubsection{Classifiers}
\label{threecls}
Generally, the ConvNet's classifier uses the dot-product operator to compute classification scores: $s = z^{T}*{w_{k}^b}$, where $z$ is the feature vector extracted by ConvNets and $w_{k}^b$ is the $k$-th classification weight vector in $W_{base}$. It is trained from scratch by thousands of optimization steps (e.g., SGD). In contrast, the $W_{base} $ is not adapted to the new categories and it is difficult to find the proper classification weights $W_{novel}$ with only a few samples and optimization steps. To address this critical problem, a classifier should be implemented to distinguish the new categories. To the best of our knowledge, current researches commonly choose one of the following classifiers to gain their best performance, i.e., SVM \cite{Chen2018}, cosine-similarity \cite{Brown} and nearest neighbor.

\textbf{SVM.} SVM classifier has achieved excellent performance for small training data in few-shot learning  \cite{Chen2018}. Essentially, unlike deep learning methods which need large-scale training data to learn generalization ability within classes, SVM is a classical transductive inference method aiming to build a model that is applicable to the problem domain.

\textbf{Nearest neighbor.} The Euclidean-based nearest neighbor method uses feature vector $z_s$ to build a prototype representation of each novel class for the few-shot learning scenario. Then it classifies the unlabeled data by calculating the distance from each query embedding point to the prototype.

\textbf{Cosine classifier.}
The cosine classifier has been well established as an effective similarity function for few-shot tasks \cite{ Vinyals2016}, which classifies samples by comparing the cosine similarity between  $z_s$ and $z_q$.


\subsection{Objective function}
The loss function is important to let the neural network generate separable representations for the unseen classes. For example, Siamese Nets \cite{Koch2015} applies contractive loss to few-shot tasks, so that neural networks can learn to distinguish similarities from dissimilarities. For fine-grained few-shot tasks, it is critical to develop an effective similarity constraint function to improve the discriminative power of the feature representations. Center loss \cite{Wen2016a}, which was first proposed for the face recognition problem, simultaneously learns a center for deep features of each class and penalizes the distances between the deep features and their corresponding class centers. Suppose there are $K$ classes for samples, $k^i$ is the category of the image $x_i$ and $z_i = f_{\varphi}(x_i)$ denotes the deep features extracted from $x_i$. Here is the formulation for center loss:

\begin{equation}\label{Center}
\mathcal{L}_c =  \frac{1}{2} \sum\limits_{i}^n  \| z_i - c_{k^i}  \|_{2}^{2}.
\end{equation}
The $c_{k} \in \mathbb{R}^K$ denotes the $k$-th class center of deep features. The formulation effectively characterizes the intra-class variations.
\begin{figure}[h]
	\begin{center}
		\includegraphics[width=0.8\linewidth]{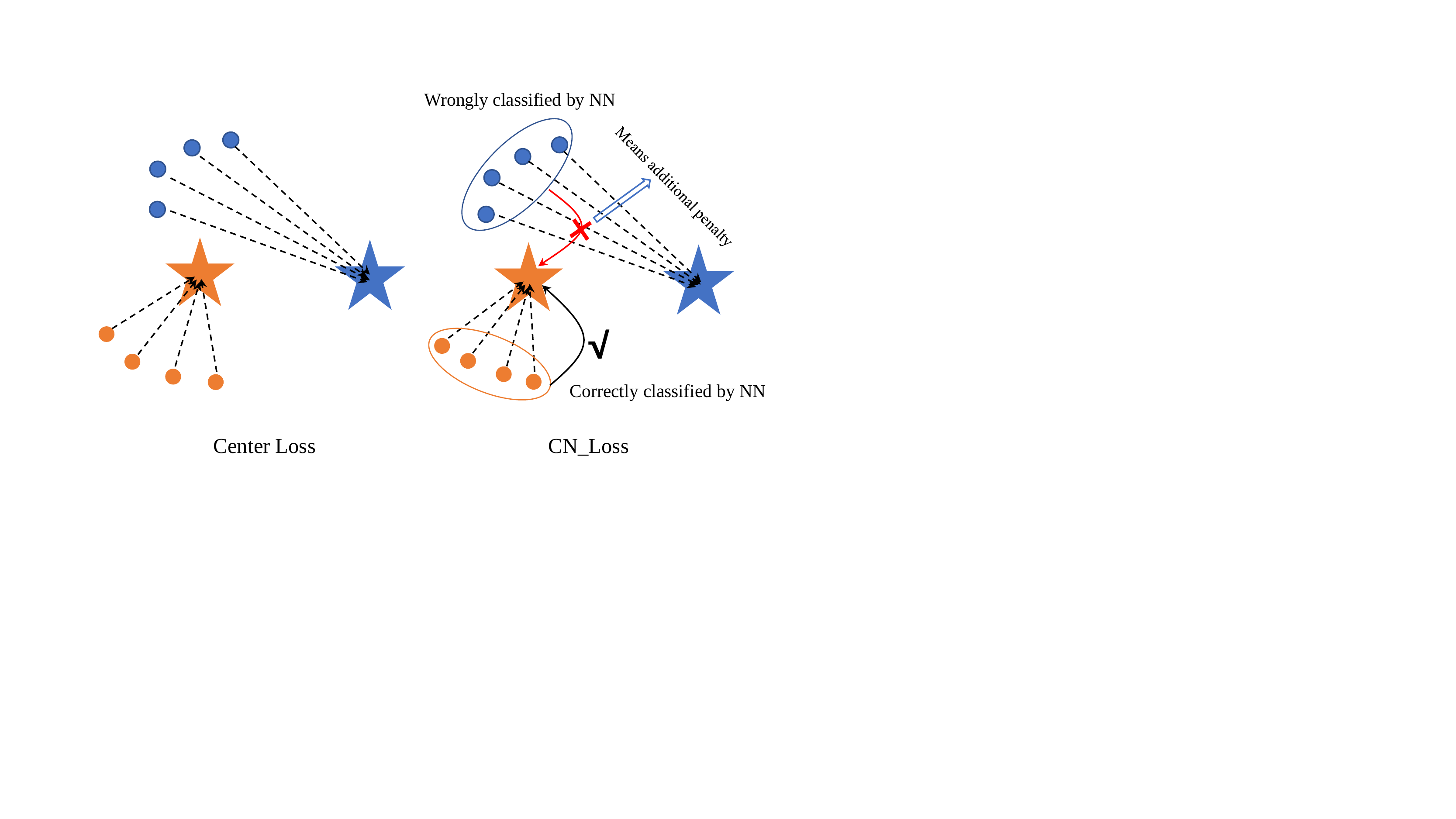}
	\end{center}
	\caption{ The center loss is simply pulling the samples into the class-center (entagram). While CN\_loss adds additional penalties to the sample of the wrong classification (red fork symbol) using nearest-neighbor. }
	\label{lossintro}
\end{figure}
However, all training samples are treated equally when a center loss function minimizes the intra-class variations, regardless of whether the sample is easy or hard to pull into the center point. 
Intuitively, for the fine-grained tasks, the difference among classes is extremely small. It is not enough to form a good distribution by simply pulling the feature vector into the class center. It is critical to impose special penalties on the samples which are difficult to approach the center during the training process. To this end, we further propose the Center Neighbor Loss (CN\_Loss) function $\mathcal{L}_s$ to form robust embedding space distribution as following.
\begin{equation}\label{CN}
\mathcal{L}_s = \mathcal{L}_c + \beta \cdot \mathcal{L}_N
\end{equation}
$\beta$ is the balance parameter for penalty term  $\mathcal{L}_N$.  $\mathcal{L}_N$ is
a negative log-probability for samples that are not classified to the correct class center. The $\mathcal {L}_N$ can be formulated as following.
\begin{equation}\label{NN}
\mathcal{L}_N = -\log \frac{\exp(-E(\bar{z^k},\mathbf{c}_k))}{\sum_{k'\in K}\exp(-E(\bar{z^{k'}}, \mathbf{c}_{k'}))}
\end{equation}
$\bar{z^k}=Avg(\sum_{x_i^k \in \mathcal{D}_{base}} f_{\varphi}(x_i^k))$ is the  $k$-th class average feature vector contained in every batch, and $E(\cdot,\cdot)$ denotes the $Euclidean$ distance.

The schematic is shown in Fig.~\ref{lossintro}. We take the center points learned from the last iteration as support points and use Euclidean-based nearest neighbors to classify the current batch of samples. With the penalty $\mathcal{L}_N$, each cluster will gather faster and perform robustly.

Ideally, the class center $c_{k}$ should be updated as feature vectors change. That means we should take the entire training set into account and average the deep features of each class in each iteration, which is not feasible in practice. To solve this problem, we implement the solution suggested for center loss \cite{Wen2016a}. First of all, we perform the update procedure based on mini-batch. The centers are computed by averaging the features of every category in each iteration. Secondly, we use the centers learned from the last iteration to classify the current batch samples by Nearest Neighbor Algorithm and punish the mislabeled samples. At last, we fix the learning rate of the centers as 0.5 to avoid large perturbations caused by mislabeled samples \cite{Wen2016a}.

\section{Experiments}
\subsection{Experimental design}
For rare categories, it's extremely difficult to collect sufficient and diverse training images. Currently, most of the previous few-shot learning methods take the {\it mini}ImageNet dataset \cite{Vinyals2016} to test their performance with 5-way 1-shot or 5-way 5-shot assumptions. However, the {\it mini}ImageNet consists of 60,000 color images with 100 classes of which 64 classes for training. The training data is enough to learn a good feature extractor for a common few-shot classification task, and nearly 80\% accuracy has been already achieved recently \cite{Chen2018}. In this paper, we focus on the fine-grained few-shot classification tasks. To this end, we design three different experiments on  Caltech-UCSD Birds  \cite{Wah2011} datasets, {\it mini}DogsNet \cite{Hilliard2018} and miniPPlankton.

For the {\it mini}DogsNet dataset \cite{Hilliard2018}, we only use \textbf{10 classes} for training, and conduct 5-way experiments with both 1-shot and 5-shot settings. We will compare our method with other well known techniques \cite{Vinyals2016}, \cite{Snell2017, Sung2017, Finn2017, Garcia2017}. All methods are also training on these 10 classes. In order to ensure the fairness of comparison, we unify the MatchingNets \cite{Vinyals2016}, PrototypicalNets \cite{Snell2017} and Imprint \cite{Brown}' feature extractor to ResNet. As the meta-learning training strategy of the Relation Networks \cite{Sung2017} and MAML \cite{Finn2017} is difficult to be trained via deep ConvNets, we keep their original network architecture.

In real-world scenarios, humans face a large number of novel categories to be recognized. 5-way experiments only for toy examples in papers. Currently, one state-of-the-art research work Imprint \cite{Brown} implemented the Caltech-UCSD Birds dataset \cite{Wah2011} for 100-way few-shot learning problem, which is much practical. Here we will carry out experiments with the same setting of Imprint \cite{Brown}. That means we investigate the accuracy on all the novel classes. As the above-mentioned methods including RelationNets \cite{Sung2017} are designed for only 5-way experiments, it is difficult to accomplish the 100-way procedure. For example, RelationNets \cite{Sung2017} requires huge GPU memory spaces for the 100-way training. Therefore, we only set the recent work Imprinted Weights \cite{Brown} as the comparison. For few-shot tasks, the Imprinted Weights \cite{Brown} described how to add a similar capability to ConvNet classifier by directly setting the weights of the final layer from novel labeled samples. Essentially, the core of Imprinted Weights method is cosine similar function. Therefore, in the following experiments, the baseline (ResNet + cosine classifier) here is the same as the Imprint.

In a real-world scenario application, for {\it{{mini}}}{{PPlankton}, we will compare our method with MatchingNets, PrototypicalNets, Relation Net, MAML and Imprint. All of above the methods are re-implemented with ResNet as the backbone feature extractor.

\subsection{Implementation details}
	\begin{figure*}[t]
		\begin{center}
			\captionsetup[subfigure]{labelformat=empty}
			
			\subfloat[\quad \quad  epoch = 20]{
				\begin{minipage}[t]{0.22\textwidth}
					\centering
					\includegraphics[height=1.1in]{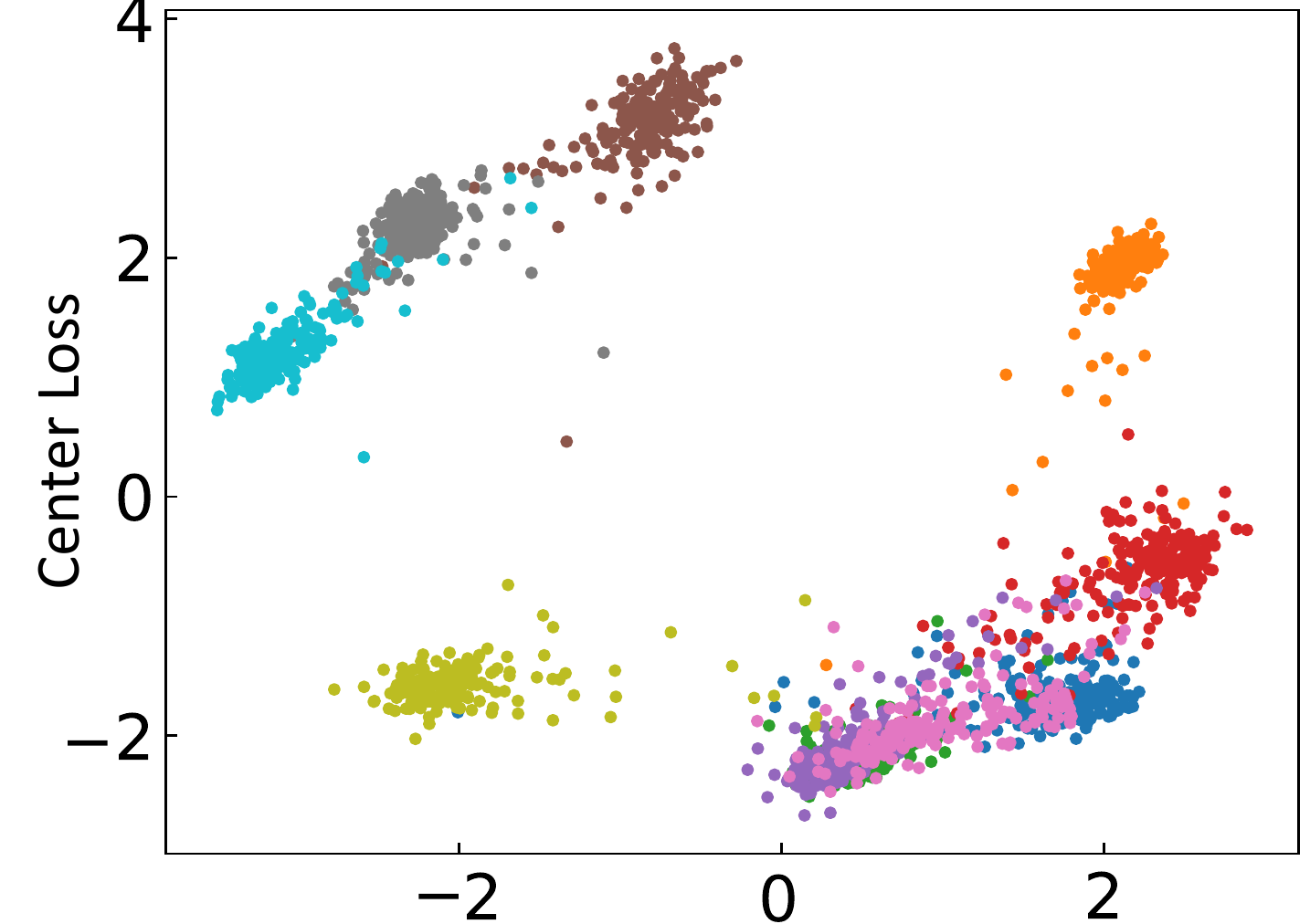} \\
					\includegraphics[height=1.1in]{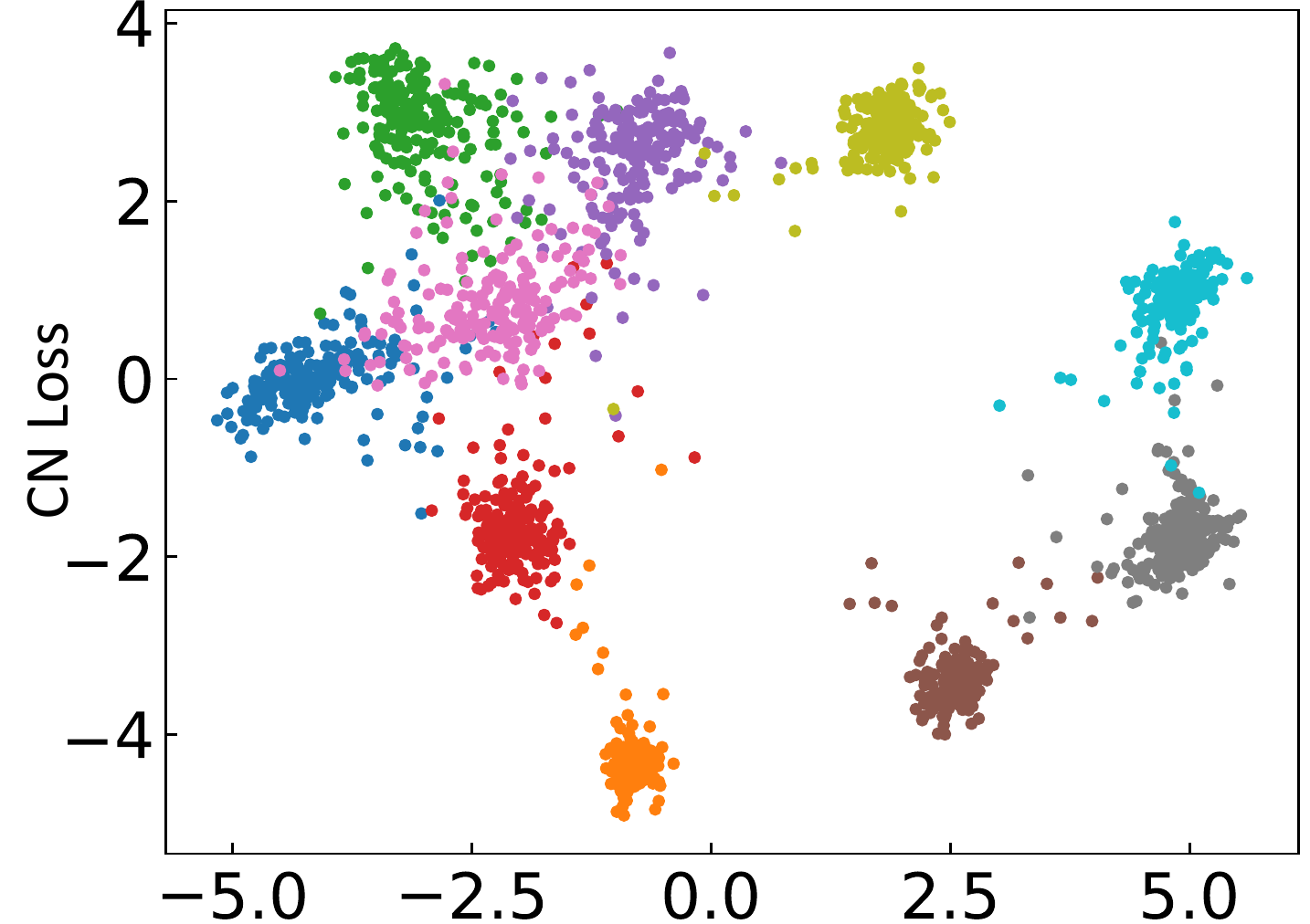}
				\end{minipage}
			}
			\subfloat[\quad epoch = 40]{
				\begin{minipage}[t]{0.22\textwidth}
					\centering
					\includegraphics[height=1.1in]{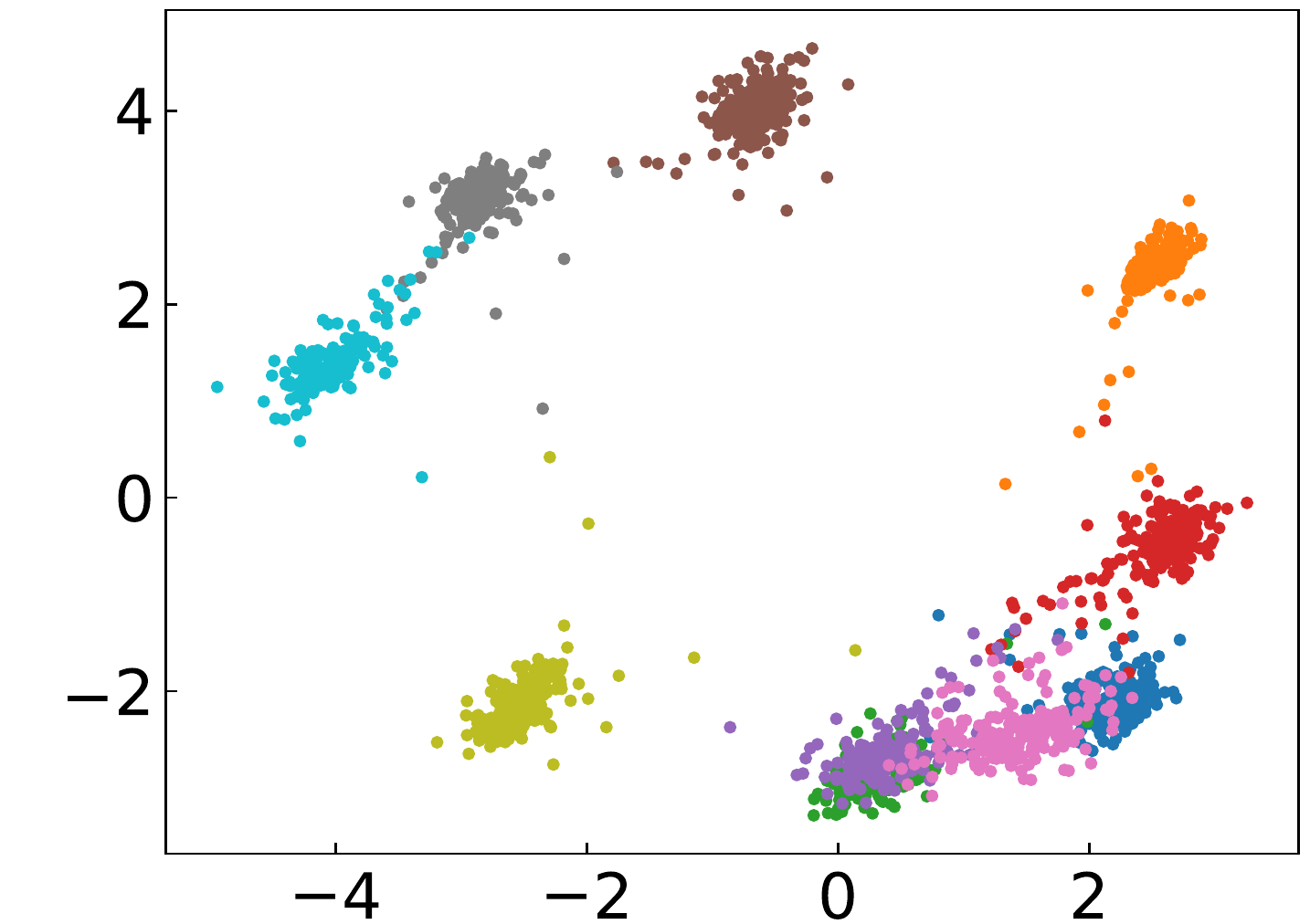} \\
					\includegraphics[height=1.1in]{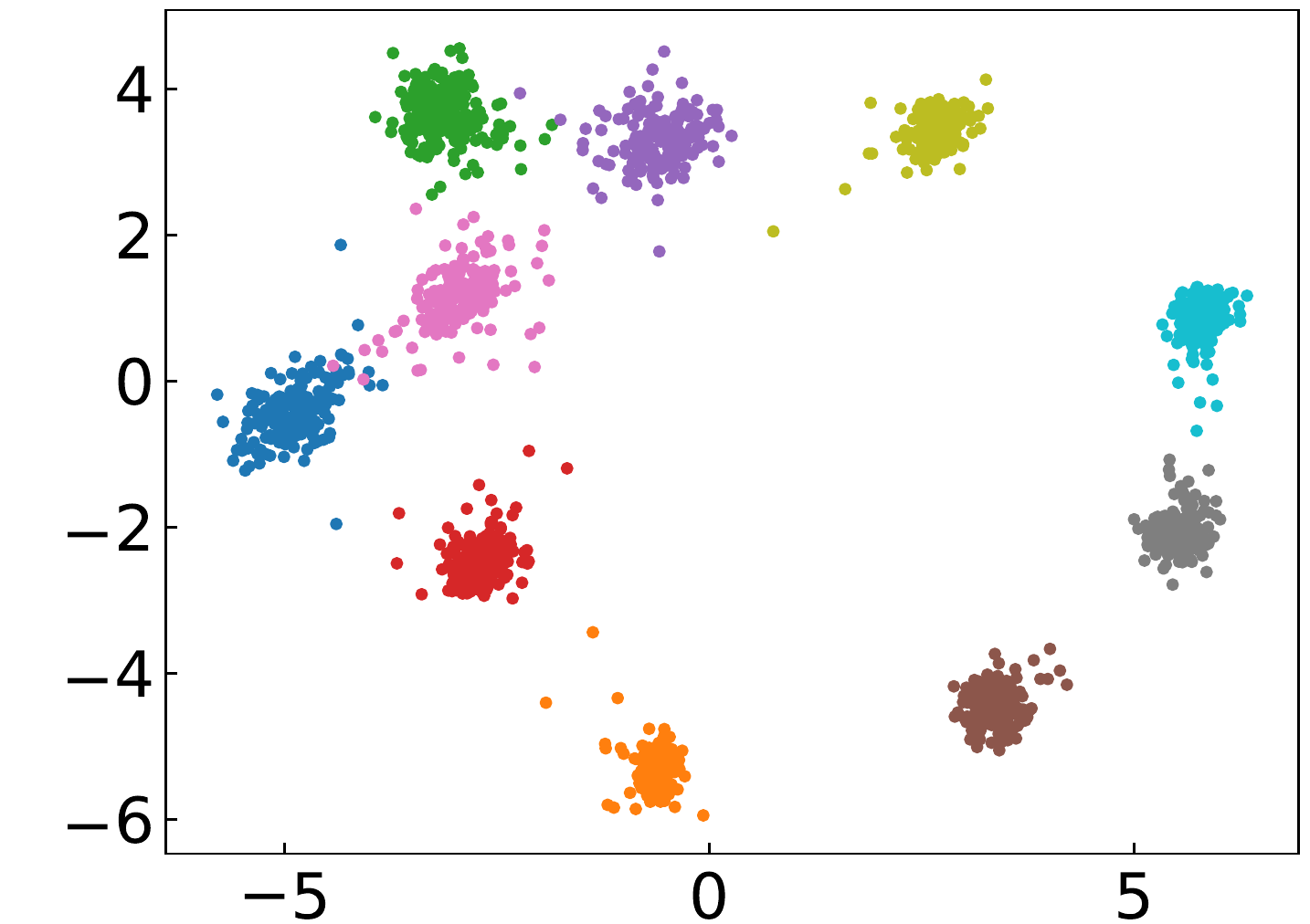} 			
				\end{minipage}
			}
			\subfloat[\quad \quad epoch = 60]{
				\begin{minipage}[t]{0.22\textwidth}
					\centering
					\includegraphics[height=1.1in]{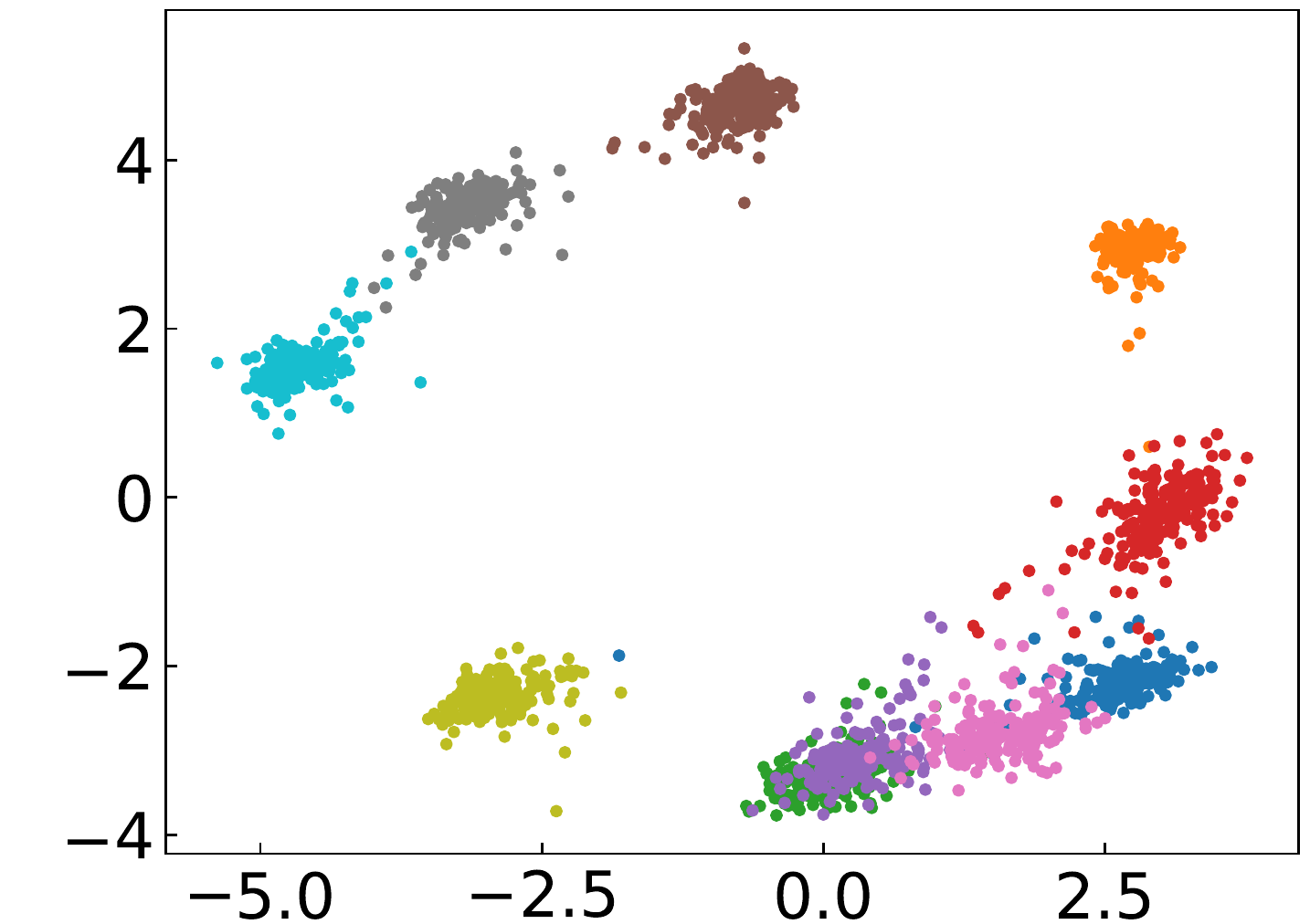} \\
					\includegraphics[height=1.1in]{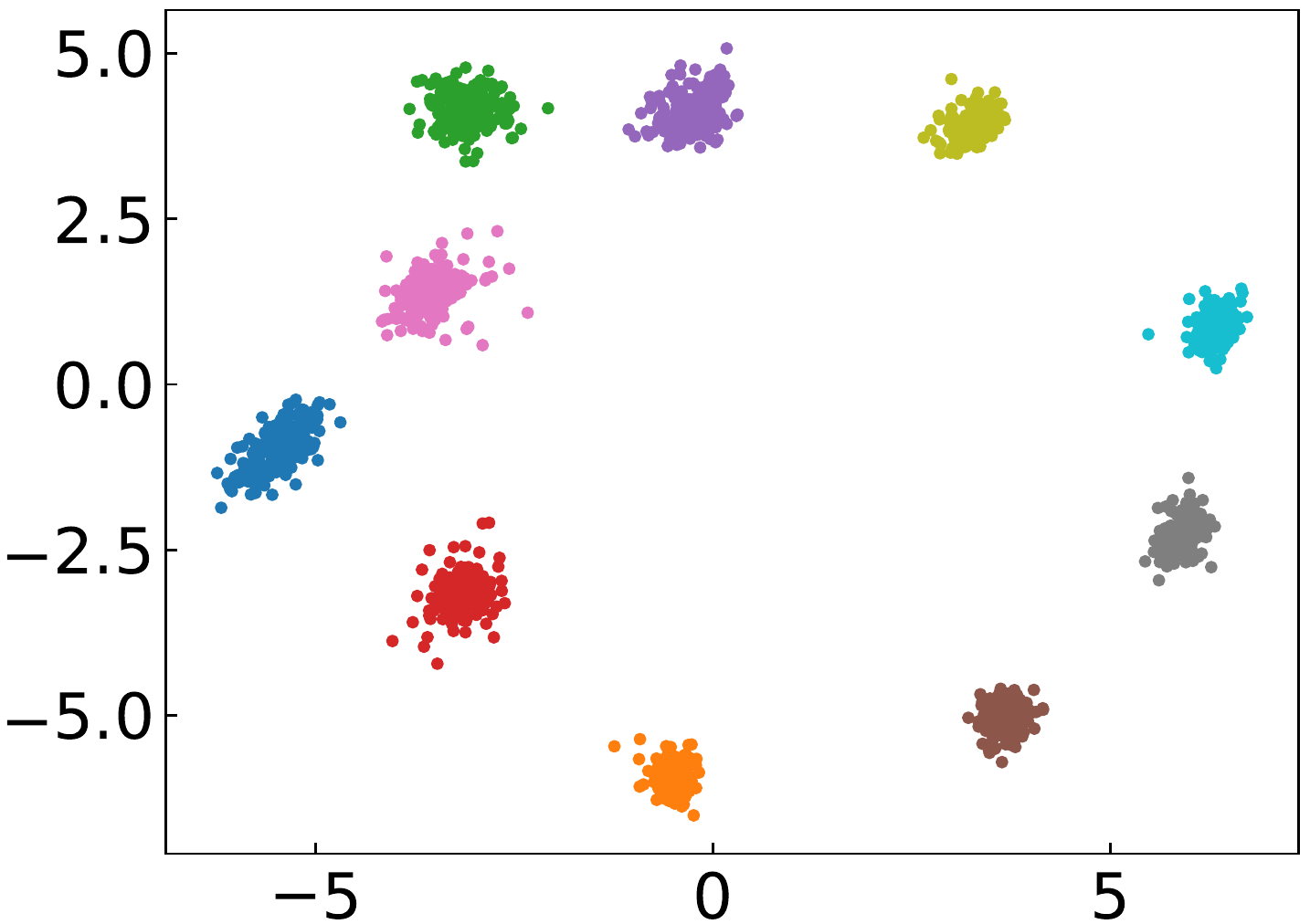} 			
				\end{minipage}
			}
			\subfloat[\quad epoch = 80]{
				\begin{minipage}[t]{0.22\textwidth}
					\centering
					\includegraphics[height=1.1in]{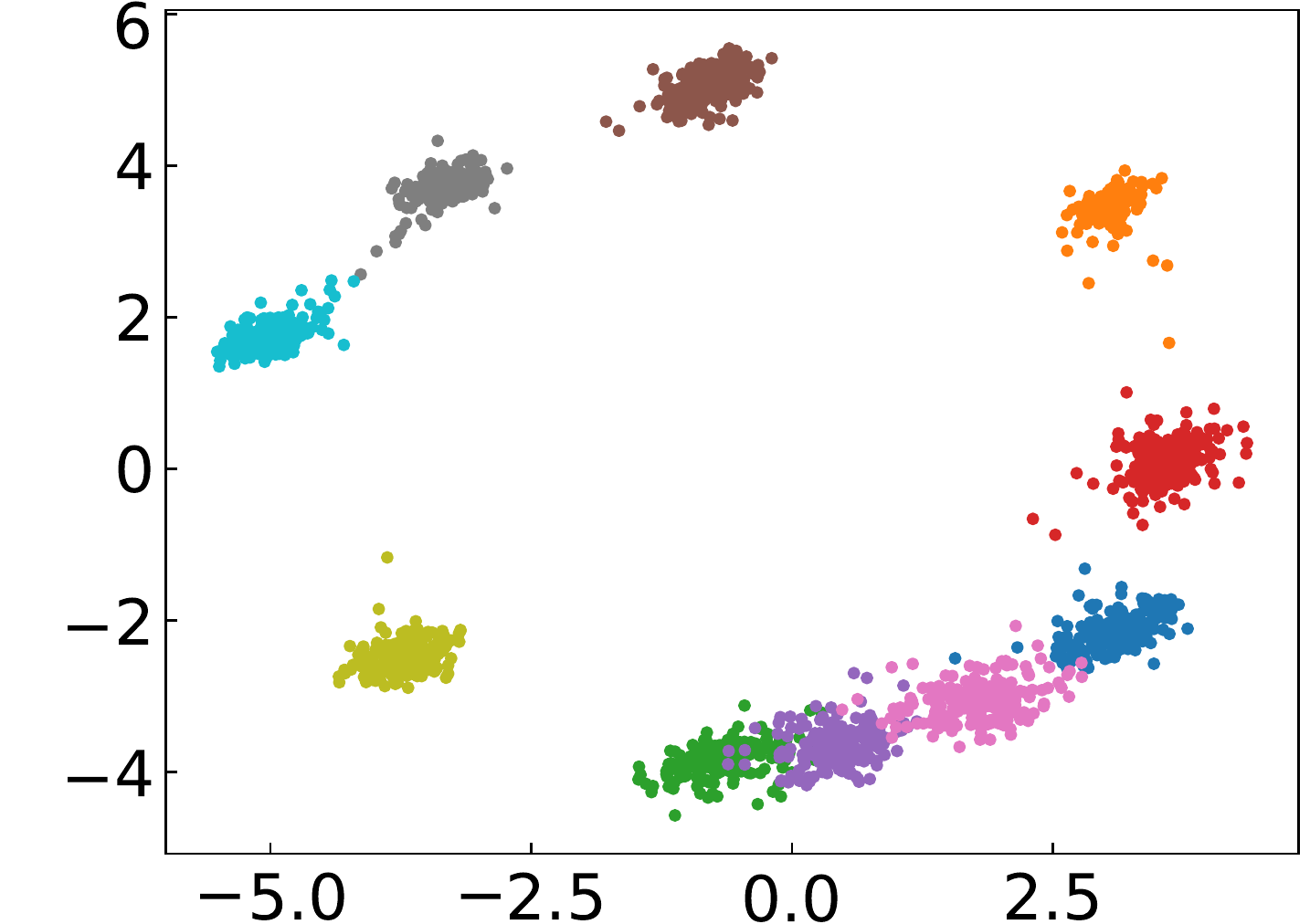} \\
					\includegraphics[height=1.1in]{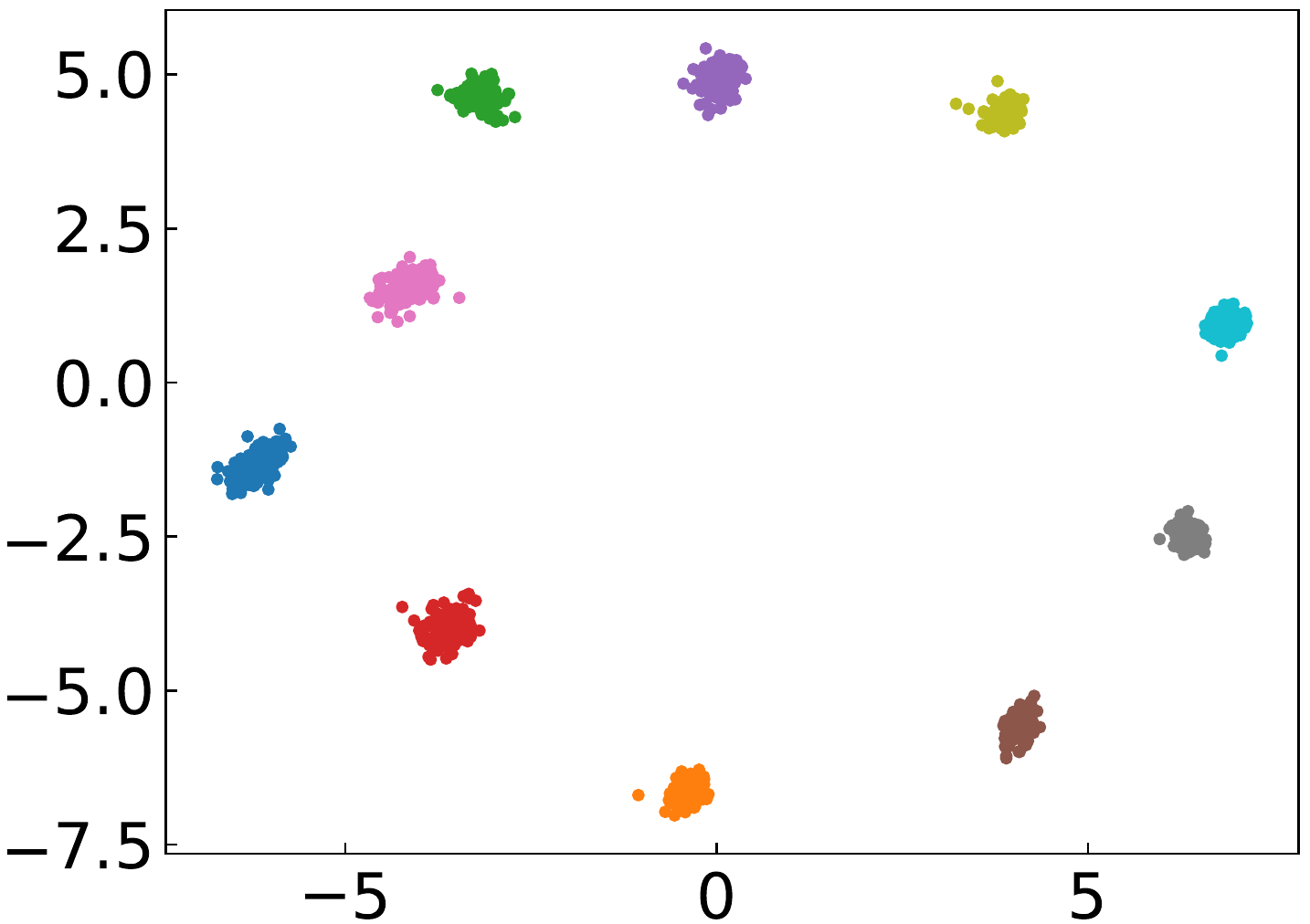} 			
				\end{minipage}
			}
		\end{center}
		\caption{The distribution of deeply learned features under Center loss and CN\_loss. Different colors denote different classes.}
		\label{distribution}
	\end{figure*}

ResNet18 \cite{He2016} is employed as the feature extractor $f_{\varphi}$. Following the similar strategy of Wen \cite{Wen2016a}, we train the feature extractor with the joint supervision of softmax loss and CN\_loss. We initialize the learning rate of the softmax loss as 0.001 and half it every 20 epochs. And we only use the last feature map as the input of high-order integration. During the testing phase, for {\it mini}DogsNet and Caltech-UCSD Birds, raw support image and zoomed focus-area are uniformly resizing into 224*224 and be sent to $f_{\varphi}$ to form a robust feature vector using element-sum operation. For {\it{{mini}}}{{PPlankton}, due to the specificity of the phytoplankton image, e.g., the target is  scattered shape. Therefore, slightly differing from the structure Fig.~\ref{fig1}, we do not use the backbone network to extract the focus-areas' feature. Here we resize focus-area into 84*84 to train a shallow CNN (four convolution blocks). Through the shallow CNN, the testing focus-area's feature  will  be concatenated with the original image's feature.

\subsection{Configuration variants}
{\bf CN\_Loss.} Fashion-MNIST is commonly used to evaluate the the loss function \cite{Wen2016a, Cale}.	We conduct similar experiments as suggested \cite{Wen2016a, Cale} to visualize the performance of Center loss and CN\_loss on Fashion-MNIST. Fashion-MNIST consists of 60,000 training examples and 10,000 for testing. Each example is a 28x28 gray-scale image, associated with a label from 10 classes. The space distribution results are shown in Fig.~\ref{distribution}. We can see that, CN\_loss can quickly form the cluster of each class.  A more robust feature space distribution usually means a better feature extractor.  And from the Table \ref{mnistacc}, the CN\_loss shows better performance on classification tasks. 
	
	\begin{table}[h]
		\small
		\begin{center}
			\begin{tabular}{cc}
				\toprule
				\bf{ Loss Function} &  \bf{Accuracy(\%)}\\
				\midrule
				Softmax Loss & {89.5 $\pm$ 0.2 } \\
				Center Loss & {90.0 $\pm$ 0.2} \\
				CN Loss &{\bf{91.42 $\pm$ 0.3}} \\
				\bottomrule
			\end{tabular}
		\end{center}
		\caption{The general classification performance of the three loss functions on the Fashion-MNIST dataset.}
		
		\label{mnistacc}		
	\end{table}

In addition, the hyperparameter $\beta$ in (\ref{CN}) is the balance for penalty term  $\mathcal{L}_N$.  We investigate the performance of our model with different hyperparameter $\beta$ on {\it mini}DogsNet's validation set. As shown in Fig.~\ref{beita}, it is very clear that the center loss (i.e., $\beta=0$) is not a good choice for few-shot classification problem. The best performance can be achieved in the case of $\beta\in [0.4,0.6]$.
	
	\begin{figure}[t]
		\begin{center}
			\includegraphics[width=0.7\linewidth]{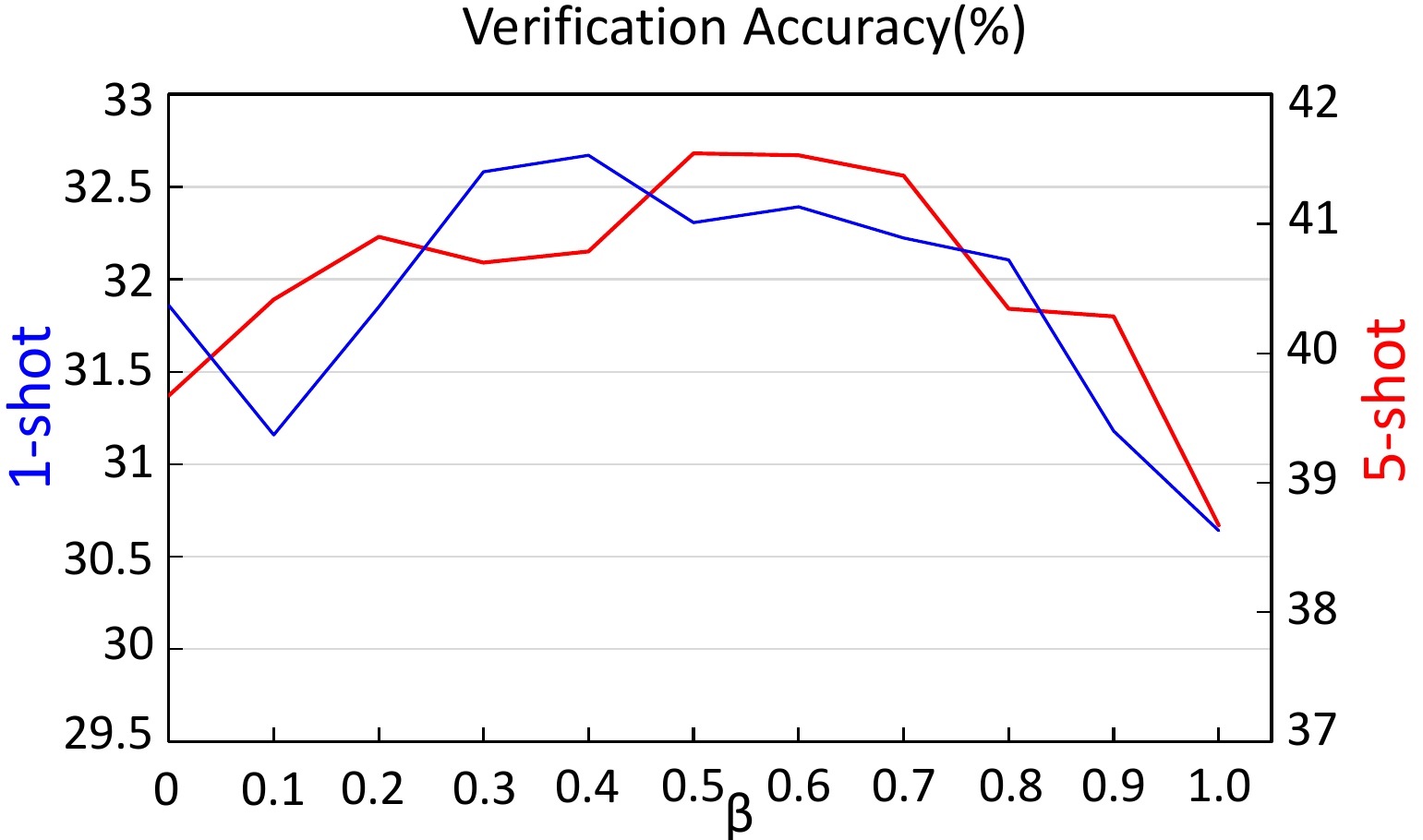}
		\end{center}
		\caption{The verification accuracy with different $\beta$.}
		\label{beita}
	\end{figure}	
	
{\bf High-order integration.} The high-order integration could help us capture more complex and high-order relationships among different intra-parts to get better attention maps. As shown in Fig.~\ref{highorder}, it helps to focus on the discriminative regions of the image. We have conducted experiments with different orders  on the performance of our method. And we found that 2-order performs stable on the novel classes classification. For instance, the accuracy of 2-order ($49.52\%$) is higher than 1-order ($48.56\%$) and 3-order ($47.00\%$) on the CUB-200-2011 dataset for 5-way setting. 

\begin{figure}[h]
	\begin{center}
		\captionsetup[subfigure]{labelformat=empty}
		\rotatebox{90}{  w/HO \quad w/o HO  \quad raw}
		\subfloat[]{
			\begin{minipage}[b]{0.06\textwidth}
				\centering
				\includegraphics[width=0.485in,height=0.485in]{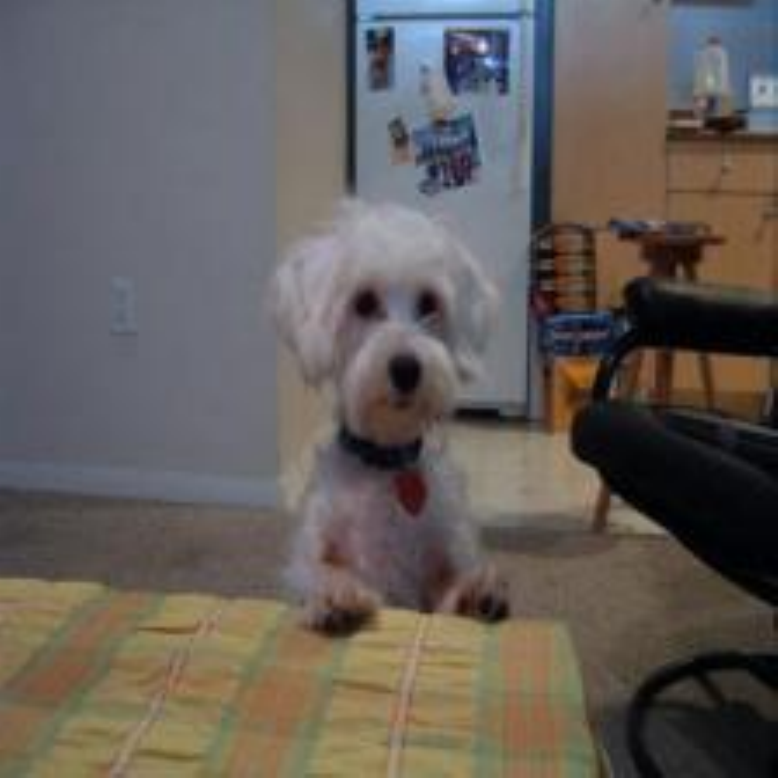} \\
				\includegraphics[width=0.485in,height=0.485in]{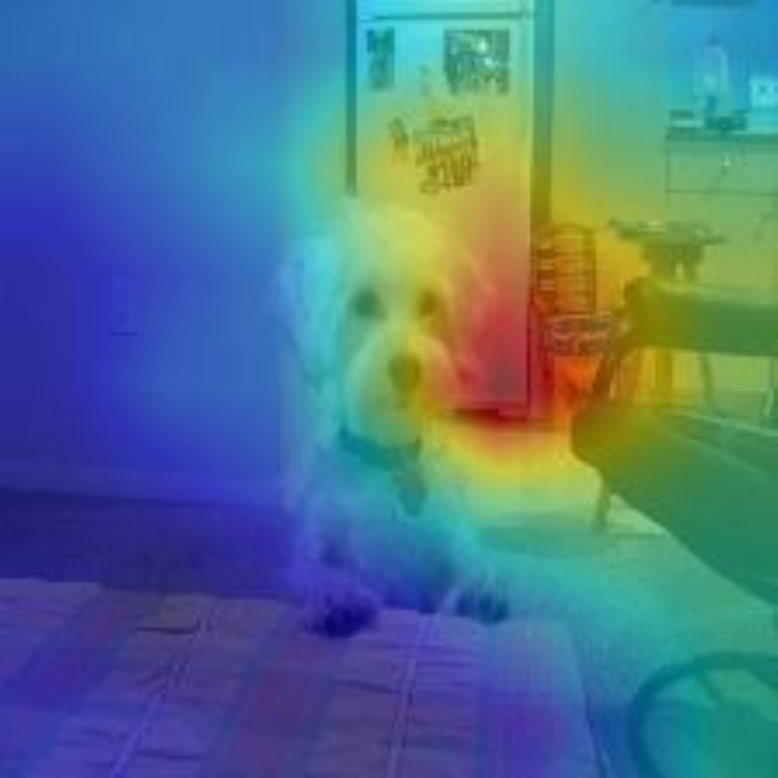} \\
				\includegraphics[width=0.485in,height=0.485in]{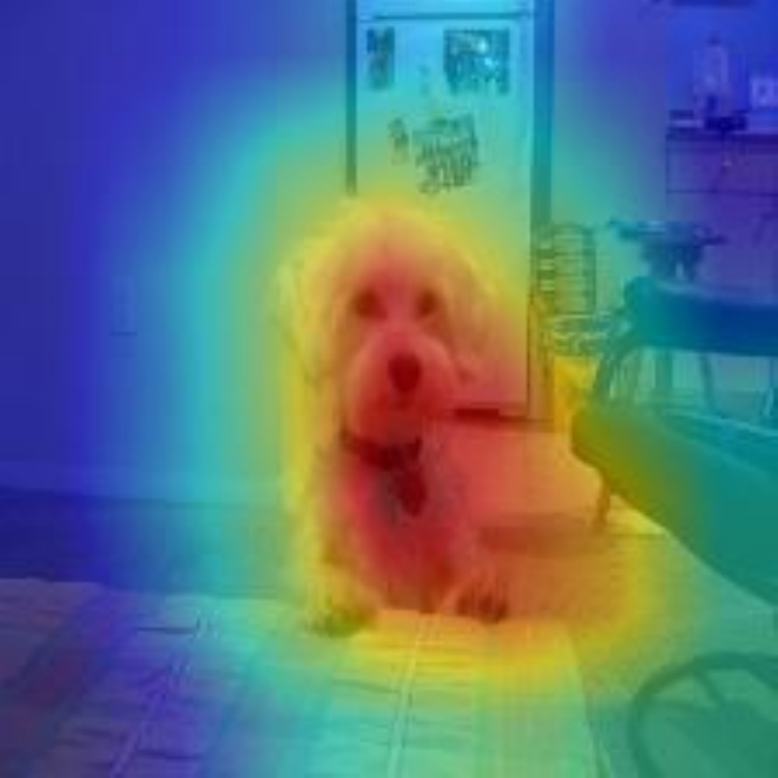} \\
			\end{minipage}
		}
		\subfloat[$\mathcal{D}_{base}$]{
			\begin{minipage}[b]{0.06\textwidth}
				\centering
				\includegraphics[width=0.485in,height=0.485in]{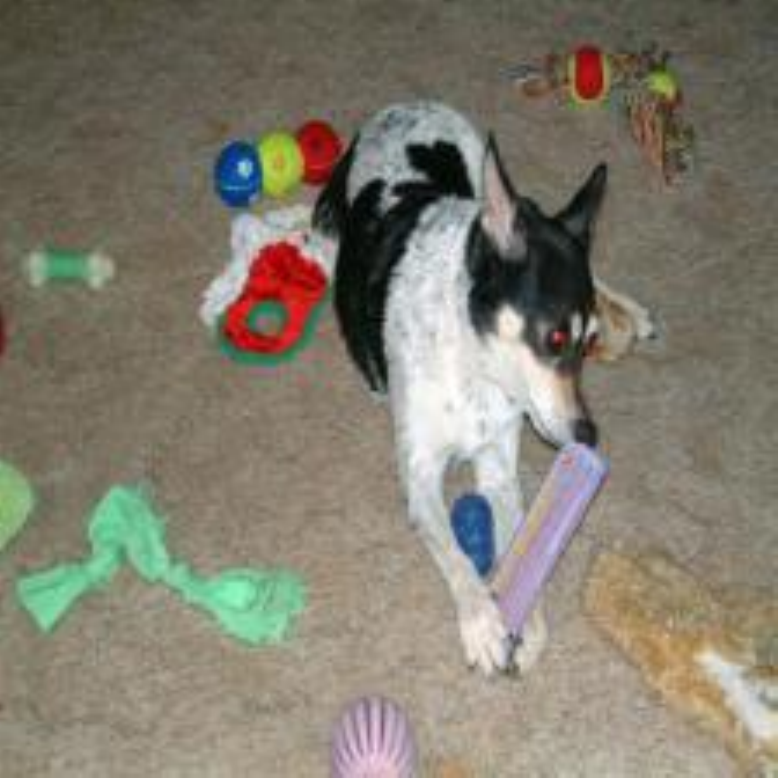} \\
				\includegraphics[width=0.485in,height=0.485in]{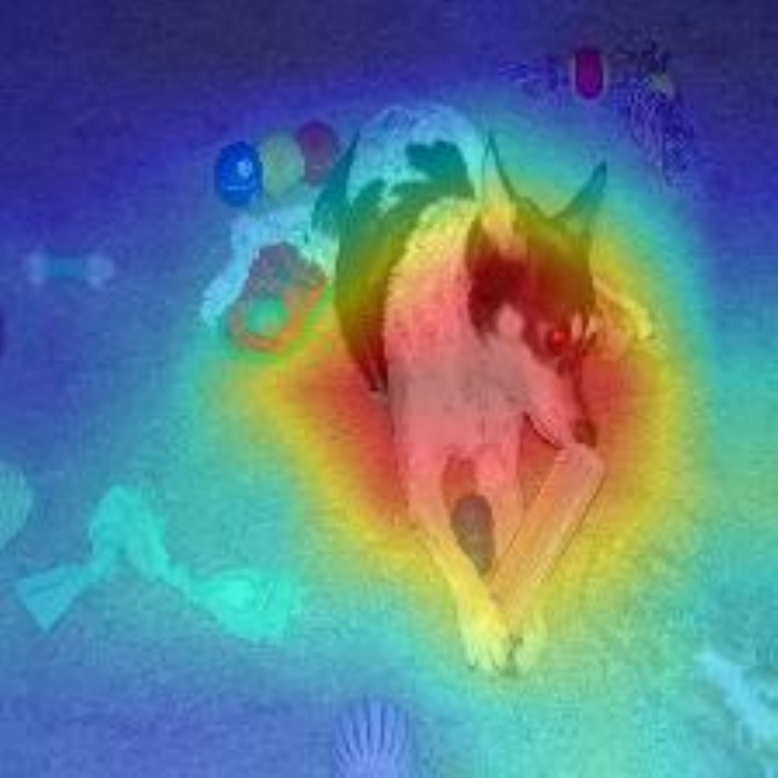} \\
				\includegraphics[width=0.485in,height=0.485in]{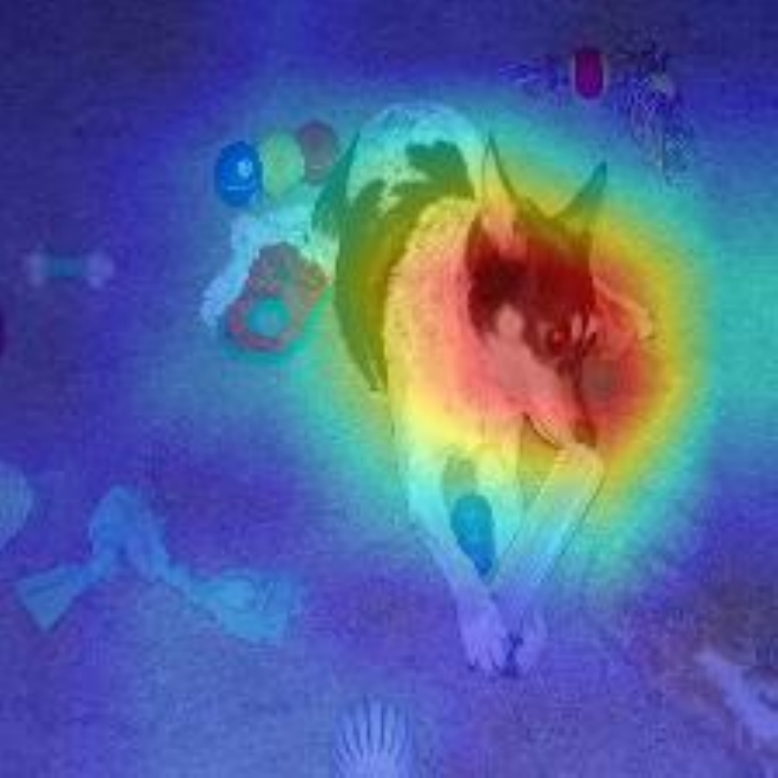} \\
			\end{minipage}
		}
		\subfloat[]{
			\begin{minipage}[b]{0.06\textwidth}
				\centering
				\includegraphics[width=0.485in,height=0.485in]{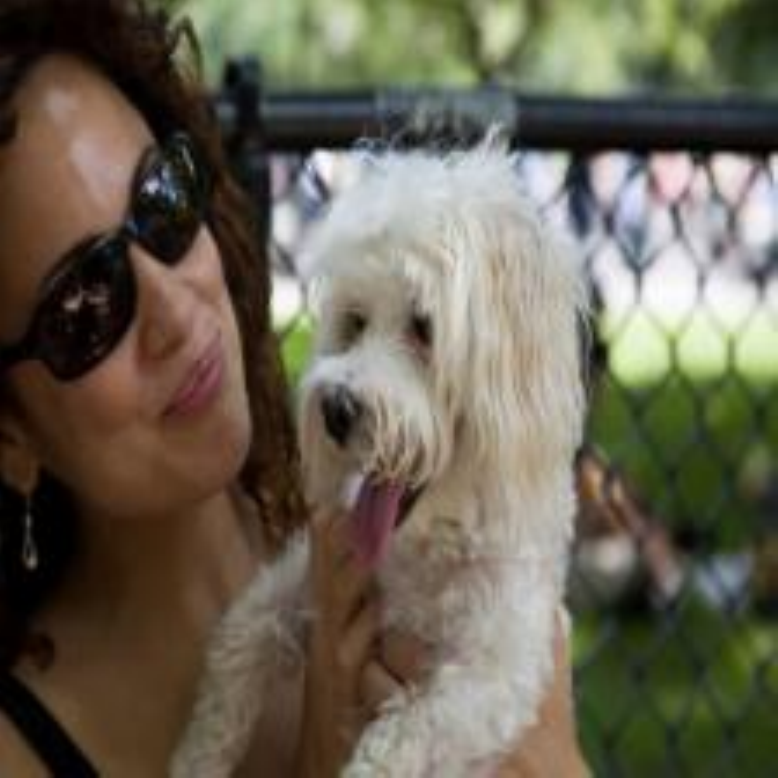} \\
				\includegraphics[width=0.485in,height=0.485in]{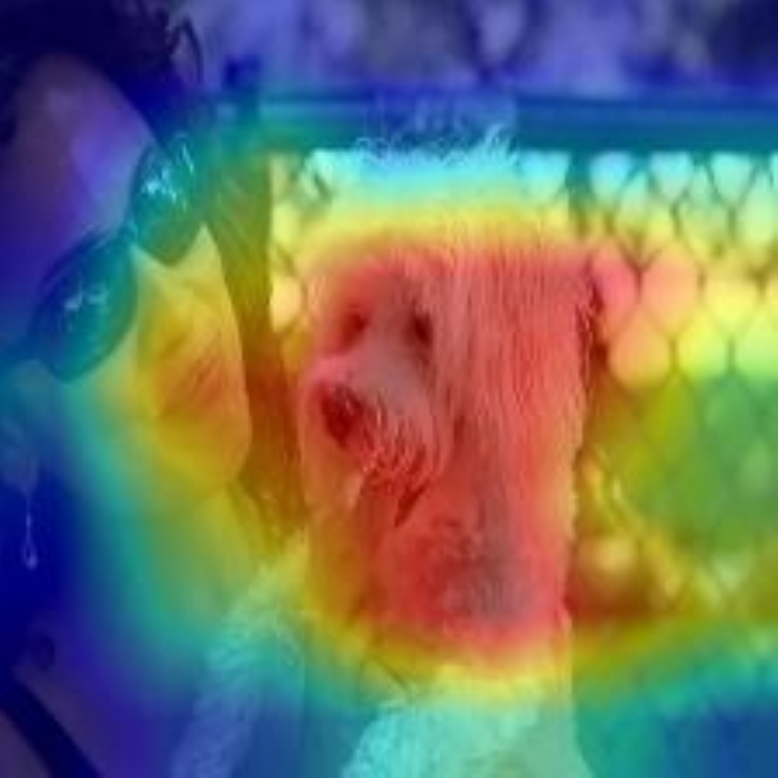} \\
				\includegraphics[width=0.485in,height=0.485in]{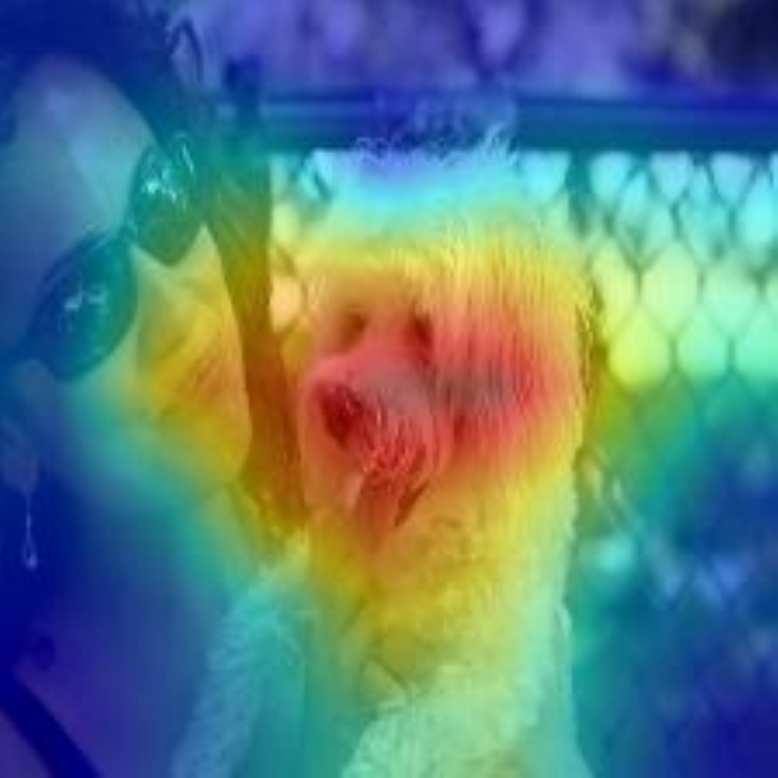} \\
			\end{minipage}
		}
		\subfloat[]{
			\begin{minipage}[b]{0.06\textwidth}
				
			\end{minipage}
		}
		\subfloat[]{
			\begin{minipage}[b]{0.06\textwidth}
				
			\end{minipage}
		}
		\subfloat[]{
			\begin{minipage}[b]{0.06\textwidth}
				\centering
				\includegraphics[width=0.48in,height=0.48in]{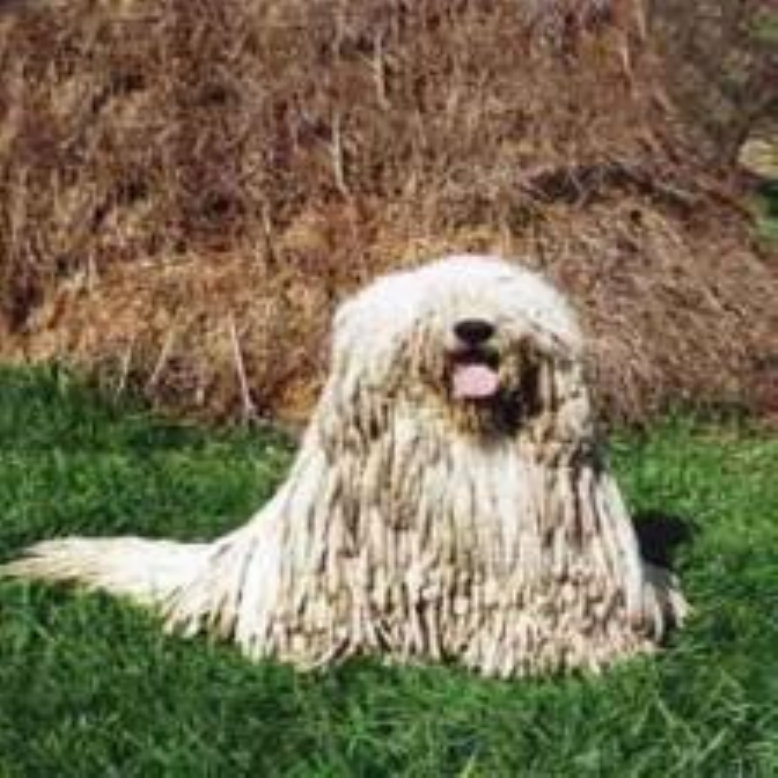} \\
				\includegraphics[width=0.48in,height=0.48in]{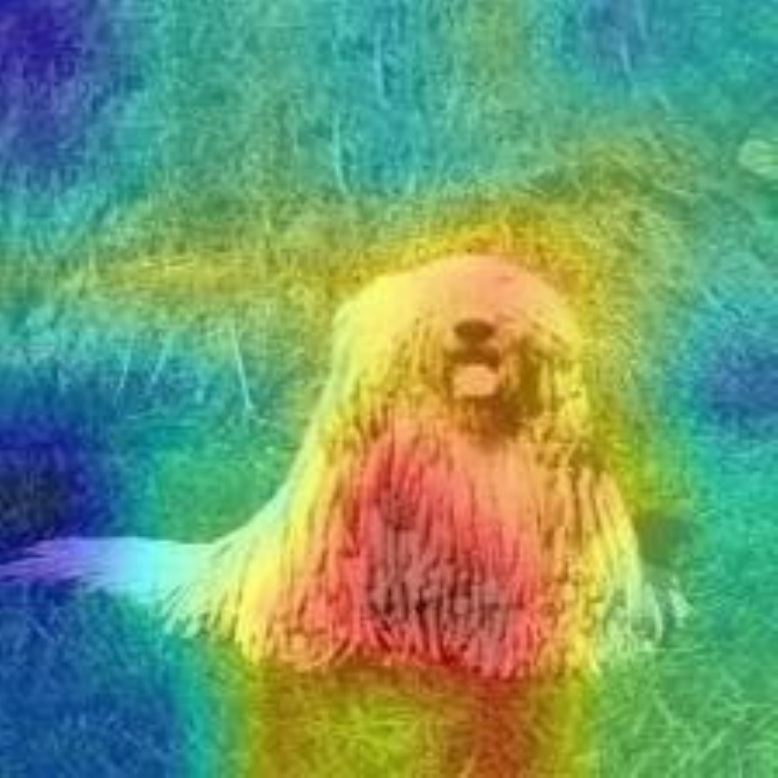} \\
				\includegraphics[width=0.48in,height=0.48in]{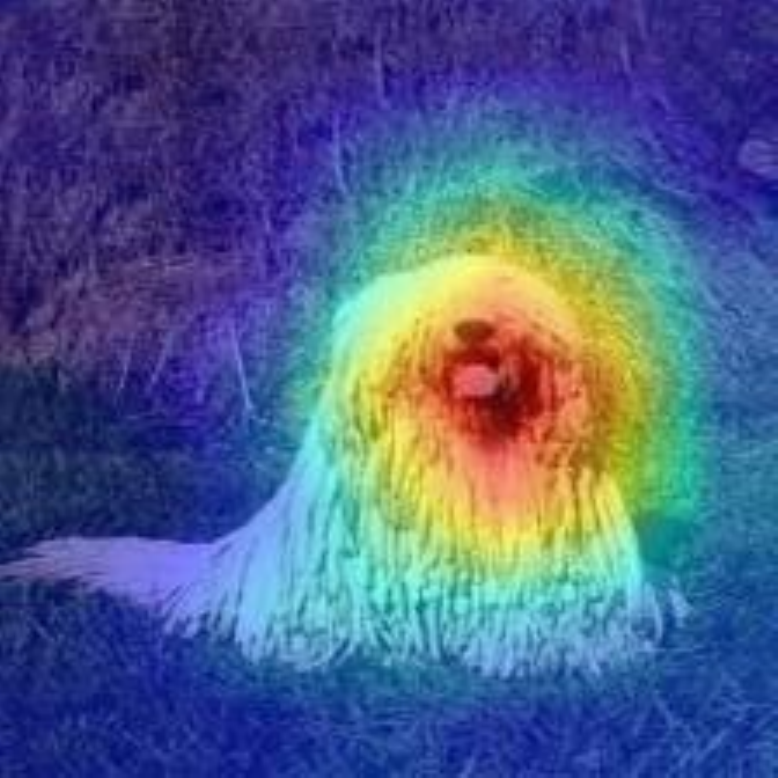} \\
			\end{minipage}
		}
		\subfloat[$\mathcal{D}_{novel}$]{
			\begin{minipage}[b]{0.06\textwidth}
				\centering
				\includegraphics[width=0.48in,height=0.48in]{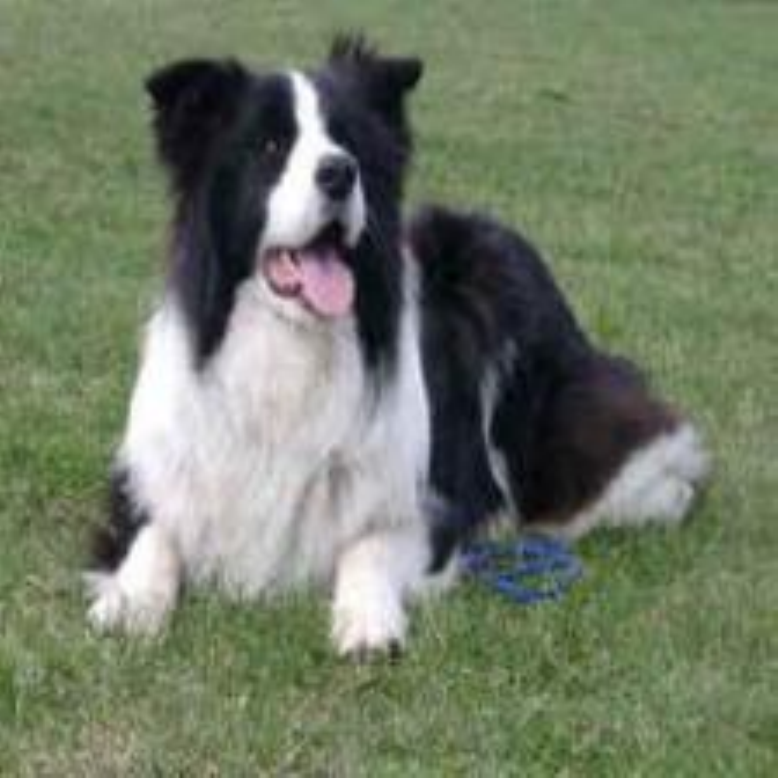} \\
				\includegraphics[width=0.48in,height=0.48in]{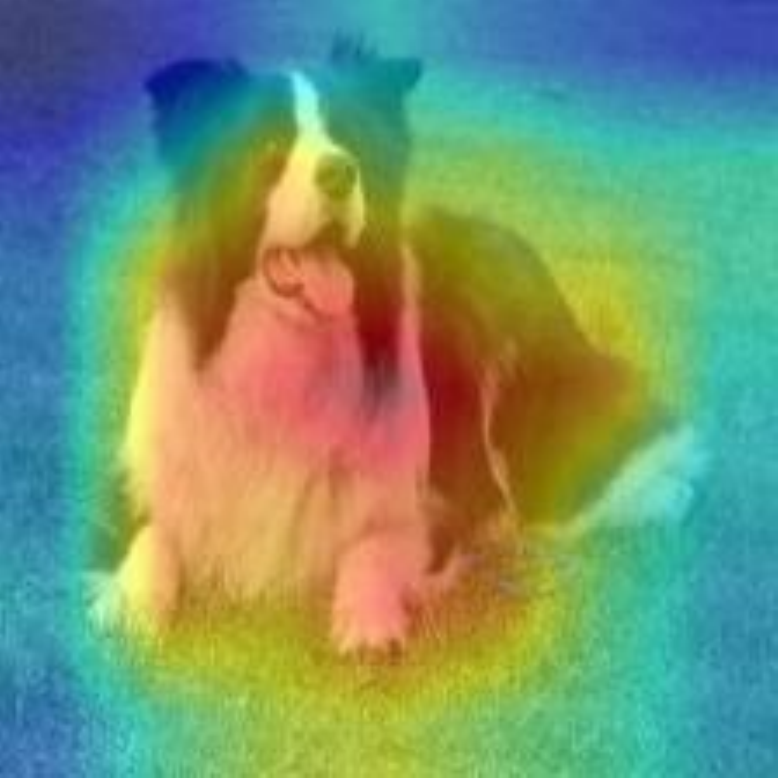} \\
				\includegraphics[width=0.48in,height=0.48in]{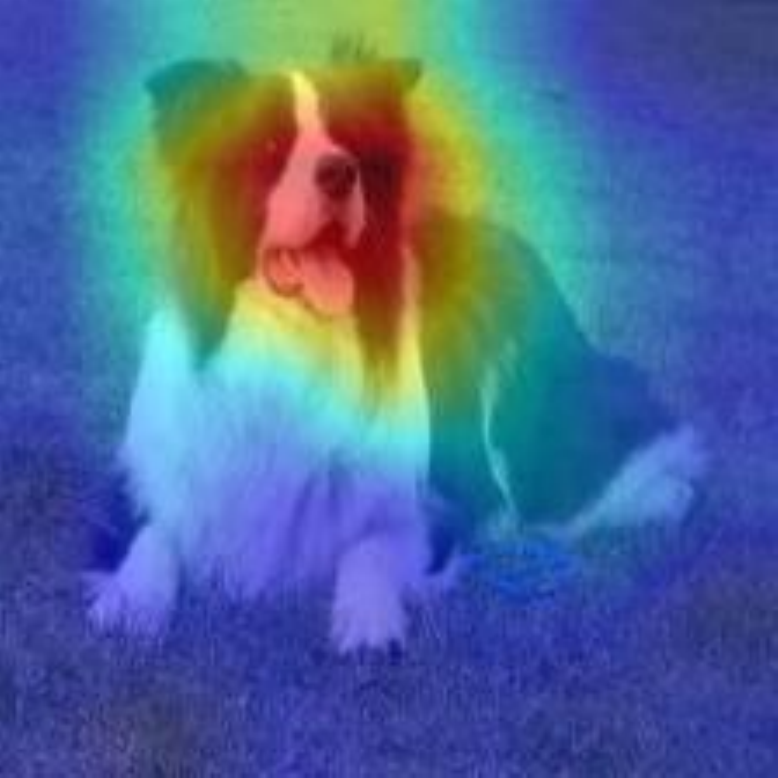} \\
			\end{minipage}
		}
		\subfloat[]{
			\begin{minipage}[b]{0.06\textwidth}
				\centering
				\includegraphics[width=0.48in,height=0.48in]{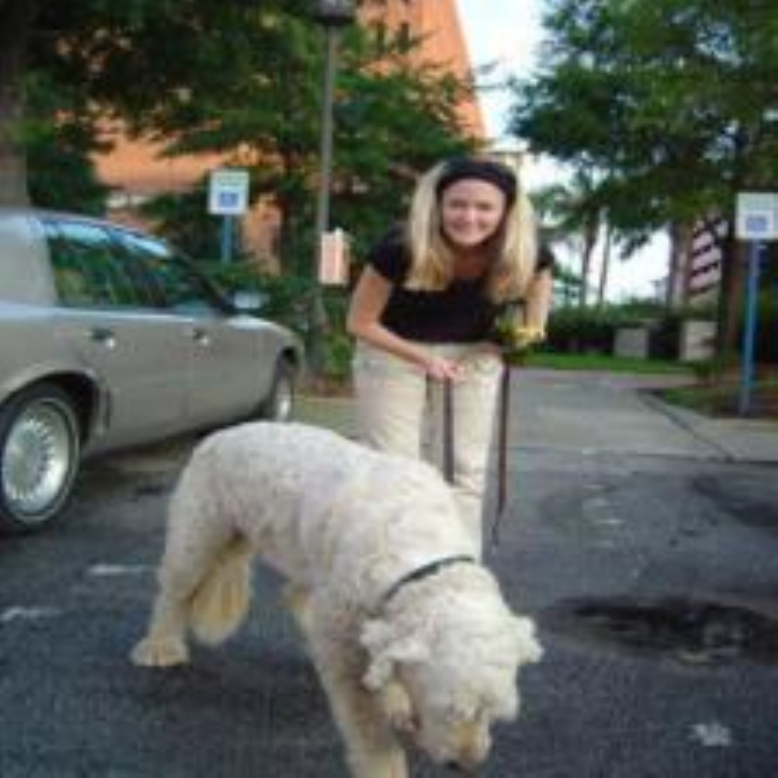} \\
				\includegraphics[width=0.48in,height=0.48in]{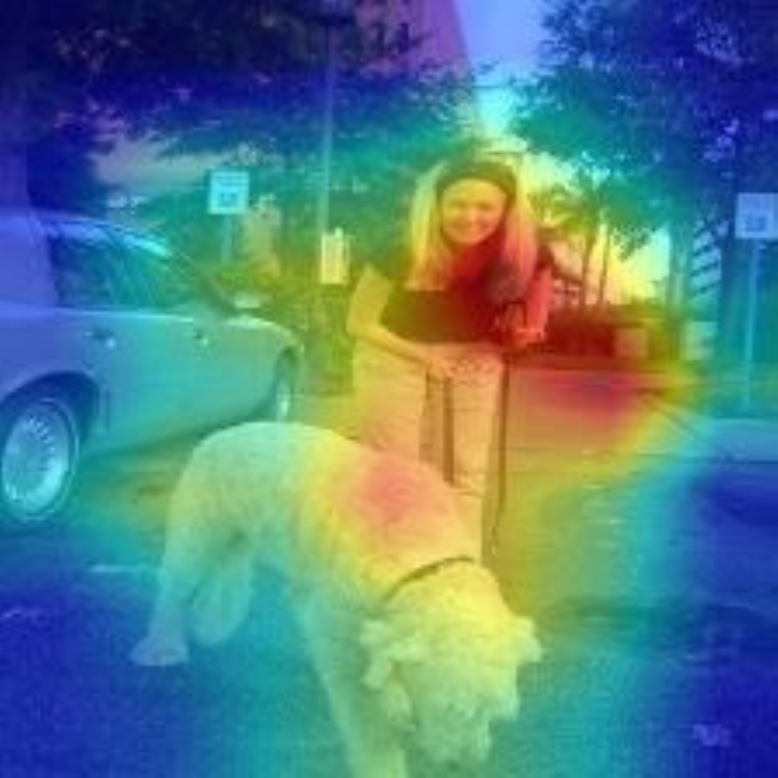} \\
				\includegraphics[width=0.48in,height=0.48in]{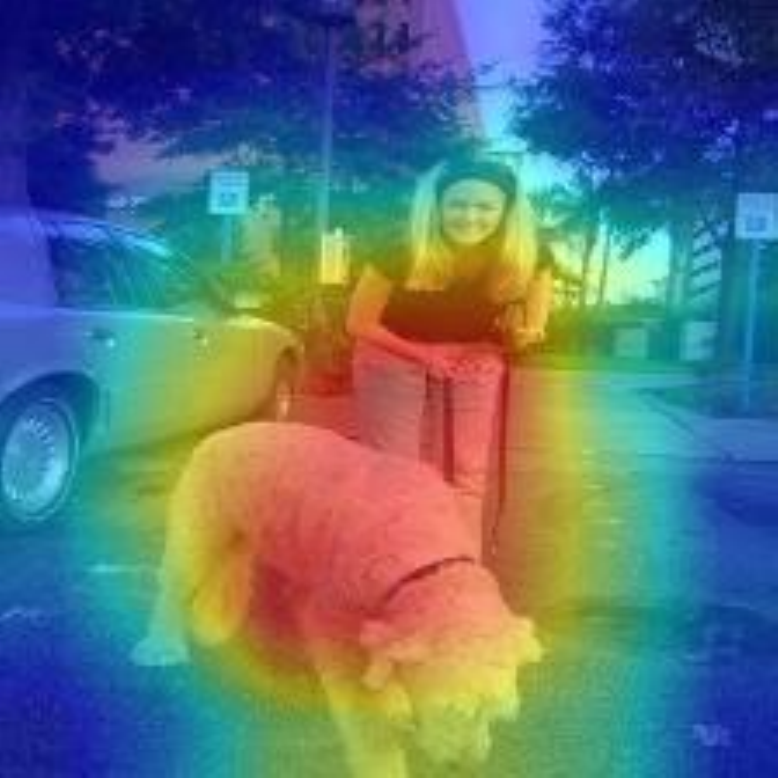} \\
			\end{minipage}
		}	
	\end{center}
	\caption{Visualization results with Higher-order Integration and without it. The left three columns show the focus regions on the $\mathcal{D}_{base}$, while the right three denote focus regions of novel samples from $\mathcal{D}_{novel}$.}\label{highorder}
\end{figure}
	
{\bf Focus-area location.} We investigate the role of Focus-area Location on our fine-grained tasks. Section~\ref{threecls} briefly described the cosine classifier used in our task. We will also illustrate the performance of different classifiers on the fine-grained few-shot classification. Figure~\ref{classfier} shows the accuracy on different tasks with the same feature extractor setting. For the {\it mini}DogsNet dataset, the cosine classifier could achieve the highest accuracy on the validation set. And focus-area location achieves positive improvement. For the Caltech-UCSD Birds, both cosine classifier and SVM can achieve nice performance. Especially, we also find that the focus-area location greatly improves the accuracy of SVM classifier.

\begin{figure}[t]
	\centering
	\captionsetup[subfigure]{labelformat=empty}
	\subfloat[]{
		\begin{minipage}[b]{0.45\columnwidth}
						\includegraphics[width=1\linewidth, height=1.32in]{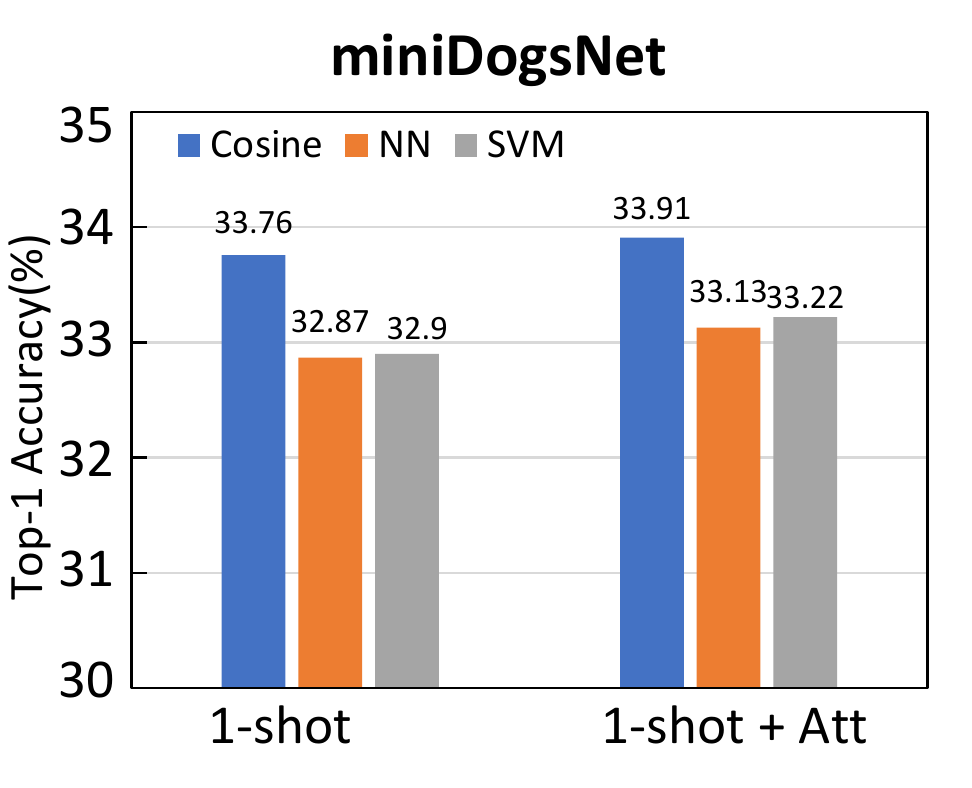}
			\includegraphics[width=1\linewidth, height=1.32in]{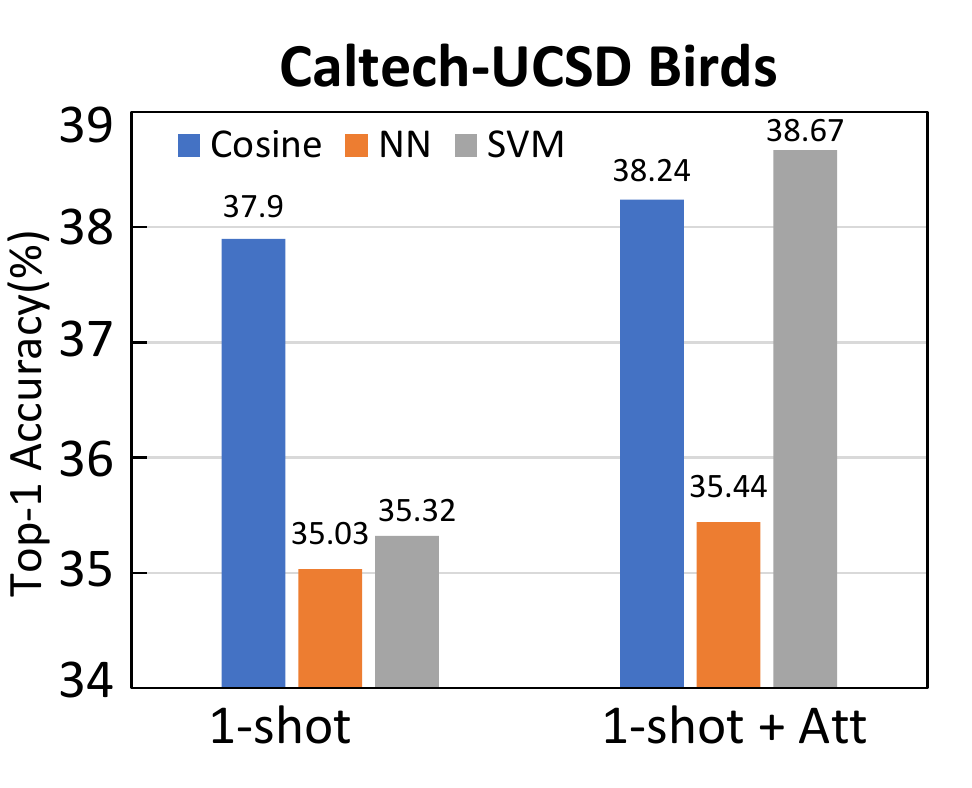}
		\end{minipage}		
	}
	\subfloat[]{
		\begin{minipage}[b]{0.45\columnwidth}
						\includegraphics[width=1\linewidth,height=1.32in]{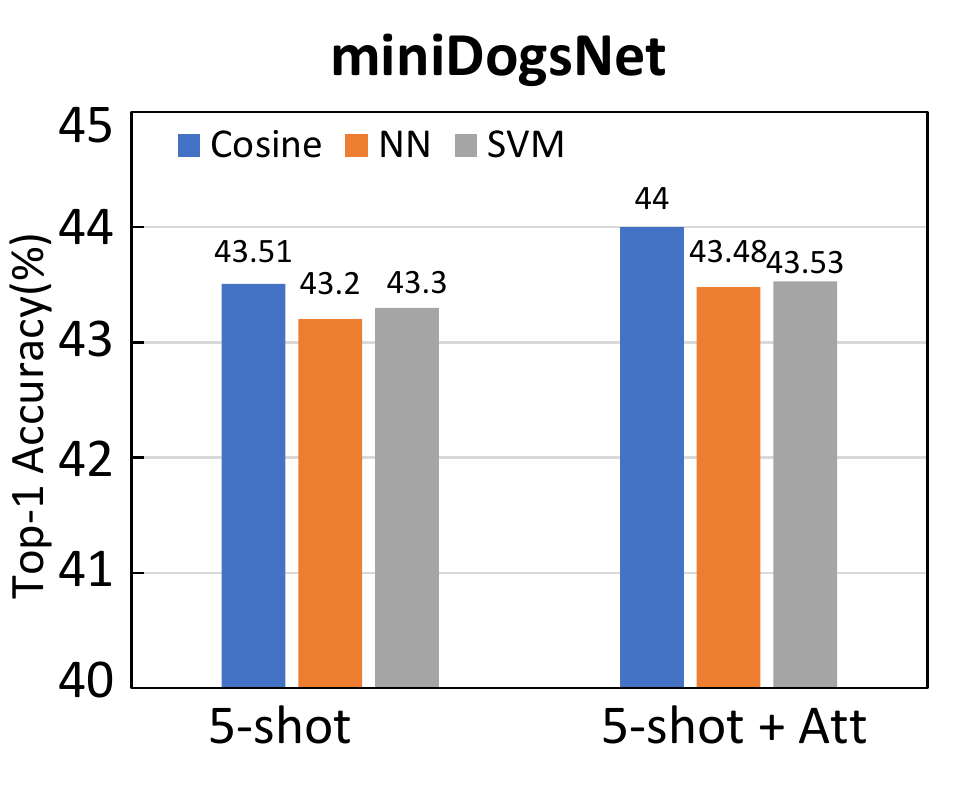}
			\includegraphics[width=1\linewidth,height=1.32in]{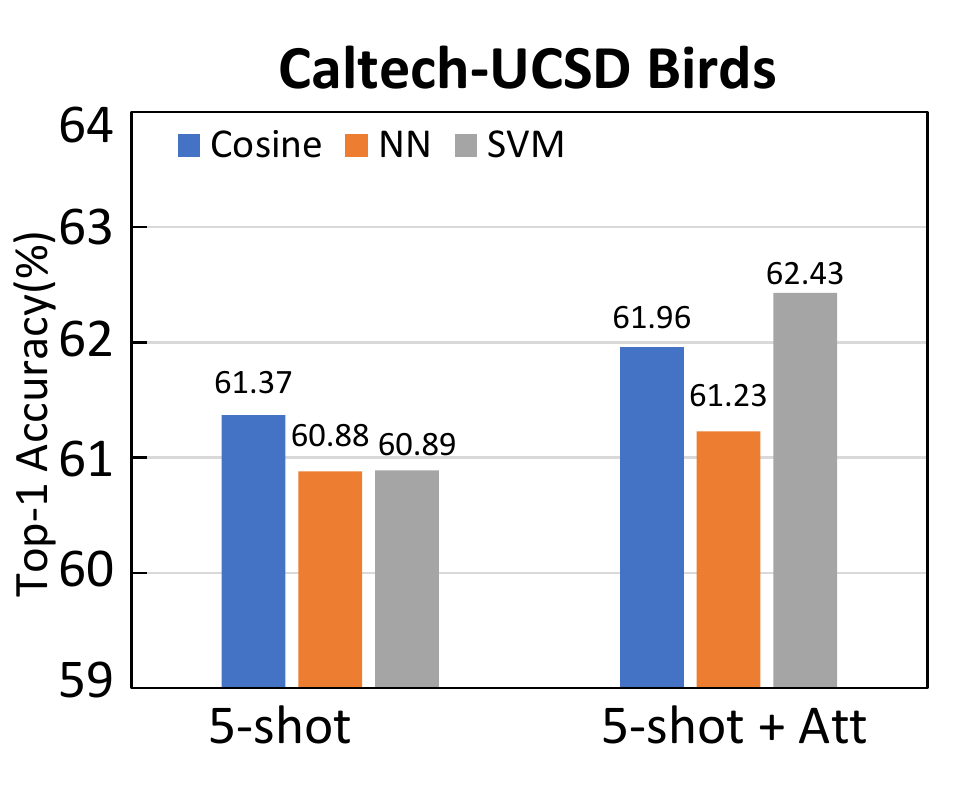}
	\end{minipage}}	
	\caption{The accuracy of three classifiers with and without Focus-area Location under 1 shot and 5 shot assumptions. }\label{classfier}
\end{figure}

\subsection{Caltech-UCSD Birds}

The Caltech-UCSD Birds dataset \cite{Wah2011} includes 200 fine-grained categories of birds with 11,788 images. We take the pre-trained ResNet18 \cite{He2016} with ImageNet as the feature extractor $f_{\varphi}$. Train/test split setting is followed the suggestion of Imprinted Weights \cite{Brown}. Here, 100 novel classes are required to be distinguished, which is very challenging and similar to the real-world scenario. The cosine classifier is employed to recognize the novel categories.

As shown in Table~\ref{tcub}, for all novel categories classification, we observe that the high-order module and CN\_Loss function are beneficial to our tasks. In particular, the information of focus-areas brings considerable improvement in accuracy on the 5-shot setting.  It is also important to illustrate the capability of recognition performance on all the categories \cite{Gidaris2018}. We further evaluate the performance on dataset of $\mathcal{D}_{base} \cup \mathcal{D}_{novel}$ \cite{Brown}. Table~\ref{tcub} and table~\ref{tcuball} show that our model achieves promising accuracies on the novel categories while at the same time it does not sacrifice the recognition performance of the base categories  $\mathcal{D}_{base}$.


\begin{table}[t]
	
	\small
	\begin{center}
		
		\begin{tabular}{cccccccc}
			\toprule
			\textbf{{$N$-shot}} &  \bf{1} & \bf{2} & \bf{5}\\
			\midrule
			Imprint (ori)\cite{Brown}&  21.26\% &  28.69\% & 39.52\% \\
			\specialrule{0em}{1.25pt}{1.25pt}
			Imprint + Aug (ori) & 21.40\% &  30.03\% & 39.35\% \\
			\specialrule{0em}{1.25pt}{1.25pt}
			Imprint (re) &28.77\% & 39.25\% &49.33\% \\
			\midrule
			ResNet + H &  30.14\% & 38.46\% & 49.83\% &  \\
			\specialrule{0em}{1.25pt}{1.25pt}
			ResNet + CNloss&  29.86\% & 39.45\% & 50.68\% &  \\		
			\specialrule{0em}{1.25pt}{1.25pt}
			ResNet + CNloss + H& 30.17\%  & 40.10\% & 50.78\%&  \\
			\specialrule{0em}{1.25pt}{1.25pt}
			ResNet + CNlos s + H + Att&  \textbf{30.82\%} & \textbf{40.85\%} & \textbf{51.95\%}&  \\
			\bottomrule
		\end{tabular}
	\end{center}
	\caption{The top-1 accuracy measured across all 100 novel classes of Caltech-UCSD Birds. 'H' denotes the High-Order Integration and 'Att' means Focus-area Location. '(ori)' means the original data in the paper and '(re)' represents the data we re-implement with ResNet as backbone.}
	\label{tcub}
	
\end{table}
\begin{table}[t]
	\small
	\begin{center}
		\begin{tabular}{ccccccccc}
			\toprule
			\bf{$N$-shot} &  \bf{1} &  \bf{2} & \bf{5} \\
			\midrule
			Imprint(ori) \cite{Brown}& 44.75\% &  48.21\% & 52.95\% \\
			\specialrule{0em}{1.25pt}{1.25pt}
			Imprint + Aug (ori)&44.60\% &  48.48\% & 52.78\% \\
			\specialrule{0em}{1.25pt}{1.25pt}
			Imprint(re)&44.68\% & 52.19\% & 59.27\% \\
			\midrule
			
			ResNet + H& {45.72\%} &   {52.64\%}& {59.96\%} \\
			\specialrule{0em}{1.25pt}{1.25pt}
			ResNet + CNloss& {45.06\%} &  51.69\% &  {58.73\%}&  \\
			\specialrule{0em}{1.25pt}{1.25pt}
			ResNet + CNloss + H& {47.23\%} &   {54.38\%}& {60.27\%} \\
			\specialrule{0em}{1.25pt}{1.25pt}
			ResNet + CNloss + H + Att& \textbf{47.89\%} &  \textbf{54.83\%} & \textbf{61.30\%} &  \\
			\bottomrule
		\end{tabular}
	\end{center}
	\caption{Top-1 accuracy measured across base plus novel categories of Caltech-UCSD Birds. }
	
	\label{tcuball}
	
\end{table}

\subsection{ {\it{\textbf{mini}}} DogsNet}

\begin{table}[h]
	
	\small
	\begin{center}
		
		\begin{tabular}{cccccccc}
			\toprule
			\specialrule{0em}{1.25pt}{1.25pt}
			\bf{5way $N$-shot} &  \bf{Dist.} & \bf{1} & \bf{5}\\
			\specialrule{0em}{1.25pt}{1.25pt}
			\midrule
			\specialrule{0em}{1.25pt}{1.25pt}
			Matching Net \cite{Vinyals2016} &  Cosine &  30.39\% & 37.97\% \\
			\specialrule{0em}{1.25pt}{1.25pt}
			Prototypical Net \cite{Snell2017} & Euclid. &  31.37\% & 39.33\% \\
			\specialrule{0em}{1.25pt}{1.25pt}
			Relation Net\cite{Sung2017} & Deep metric & 32.42\% & 38.53\% \\
			\specialrule{0em}{1.25pt}{1.25pt}
			MAML \cite{Finn2017} & - & 26.66\% & 35.60\% \\			
			\specialrule{0em}{1.25pt}{1.25pt}
			Imprint \cite{Brown} &  Cosine & 30.14\% & 38.31\% &  \\
			\midrule
			
			\specialrule{0em}{1.25pt}{1.25pt}
			Resnet + H &  Cosine & 30.17\% & 38.77\% &  \\
			\specialrule{0em}{1.25pt}{1.25pt}
			ResNet + CNloss&  Cosine & 31.25\% & 40.63\% &  \\			
			\specialrule{0em}{1.25pt}{1.25pt}
			ResNet + CNloss + H& Cosine  & 31.95\% &41.40\%&  \\
			\specialrule{0em}{1.25pt}{1.25pt}
			ResNet+CNloss+H+Att&  Cosine & \textbf{33.13\%} & \textbf{42.53\%}&  \\
			\bottomrule
		\end{tabular}
	\end{center}
	\caption{The top-1 accuracy on the test set of {\it mini}DogsNet, all accuracy results are averaged over 100 test episodes and each episode contains 100 query samples from 5 classes. All results are reported with 95\% confidence intervals.  }
	\label{tab:mini}
\end{table}

Hilliard et al. \cite{Hilliard2018} created a {\it mini}DogsNet which consists images of dog categories from the ImageNet to test the model's fine-grained ability. They selected 100 of those classes and use the 64/16/20 random classes split for training, validation, and testing. In our work, we further increase the difficulty by random selecting 10 of 64 classes to form our training set. That means only 10 classes are used for training the feature extractor and 20 novel classes should be distinguished. And we train the ResNet18 \cite{He2016} from scratch.

\begin{figure}[t]
	\begin{center}
		\includegraphics[width=0.9\linewidth]{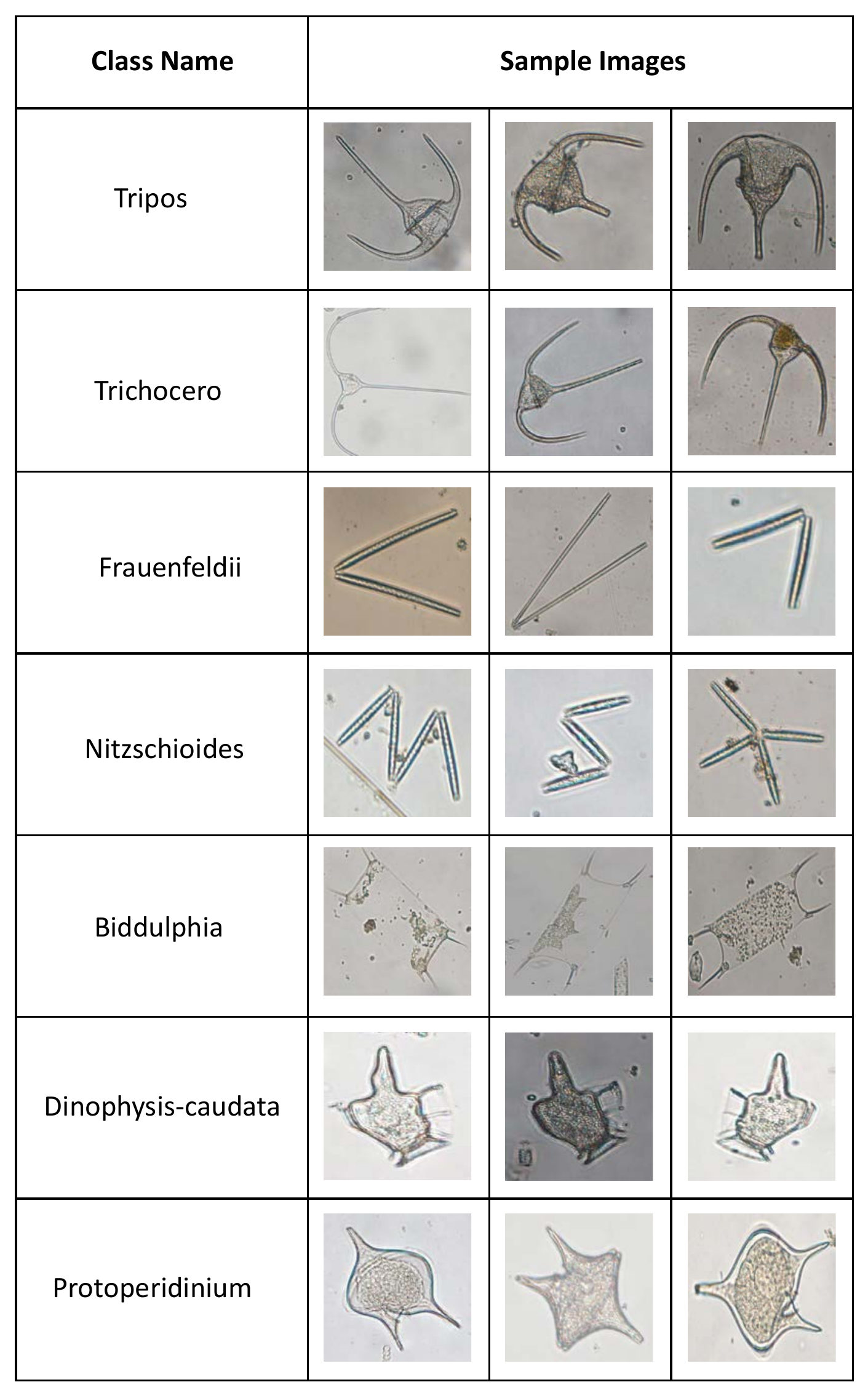}
	\end{center}
	\caption{Random samples of nine categories from our Phytoplankton dataset. The morphological differences among different categories are very small (such as the first two categories). It is a typical dataset for fine-grained challenge problem.}
	\label{plankton}
\end{figure}

We conduct 5-way experiments with both 1-shot and 5-shot trials. Table~\ref{tab:mini} shows that our model could achieve promising performance both on 1-shot and 5-shot tasks. To verify the effectiveness of our different modules, we use the ResNet18 and cosine classifier as the baseline. To our surprise, the baseline can also achieve nice performance. Relation Nets uses the deep non-linear metric to capture the similarity between samples and is well performed even using 10 classes training data. For our method, we can see that CN\_Loss and high-order integration can bring promising improvements. And the focus-area location mechanism is still beneficial to the task.

\subsection{{\it{\textbf{mini}}} PPlankton}

For a real-world task in specific domain such as phytoplankton classification, it is infeasible to collect large-scale samples and it always requires experts to label the data. Meanwhile, it is also quite difficult to search for the relevant open-source web-data. Current monitoring systems (e.g. ZooScan and FlowCAM \cite{Gorsky2010, Jakobsen2011}) yield large amounts of images every day. It requires many marine biologists to manual classify the sample images. Nevertheless, new and scarce categories are valuable for marine science. {\textbf{PPlankton} is a large-scale public dataset for machine learning with the help of marine biologists \cite{Qiong2019}. And for few-shot tasks, we further construct a phytoplankton dataset  {\it{\textbf{mini}}}{\textbf{PPlankton}. It is a particular image dataset for few-shot fine-grained classification problem.
	
Some examples of the dataset are shown in Fig.~\ref{plankton}. To construct the dataset, we collect seawater samples from the Bohai Sea, and we photograph phytoplankton images contained in the sampled seawater by optical microscopes. With the help of  marine biologists,
we label each object with its confident catergory. The {\it mini}{PPlankton} includes 20 classes each of which contains about 70 samples. From  Fig.~\ref{plankton}, we can observe that our dataset faces the challenge problem of fine-grained classification. For example, their shapes between different categories are similar, such as tripos and trichocero.

\begin{table}[h]
	\small
	\begin{center}
		\begin{tabular}{cccccccc}
			\toprule
			\specialrule{0em}{1.25pt}{1.25pt}
			\bf{$N$-shot} &  \bf{Dist.} & \bf{1} & \bf{5}\\
			\midrule
			\specialrule{0em}{1.25pt}{1.25pt}
			Matching Net \cite{Vinyals2016}& Cosine. &  48.76\% & 60.78\%&  \\
			\specialrule{0em}{1.25pt}{1.25pt}
			Prototypical Net \cite{Snell2017} & Euclid.  &  50.84\% & 66.67\%&  \\
			\specialrule{0em}{1.25pt}{1.25pt}
			Relation Net \cite{Sung2017}& Deep-metric & 46.79\% & 58.48\% &  \\
			\specialrule{0em}{1.25pt}{1.25pt}
			MAML\cite{Finn2017} & - & 46.0\% & 60.63\% &  \\	
			\specialrule{0em}{1.25pt}{1.25pt}
			Imprint\cite{Brown} & Cosine  &  57.72\% & 72.99\%&  \\
			\midrule
			\specialrule{0em}{1.25pt}{1.25pt}
			ResNet + CNloss & Cosine & 59.0\% & 74.84\% &  \\
			\specialrule{0em}{1.25pt}{1.25pt}
			ResNet + CNloss + H & Cosine  &  56.29\% & 70.8\%&  \\
			\specialrule{0em}{1.25pt}{1.25pt}
			ResNet + CNloss + Att& Cosine & \textbf{60.03\%} & \textbf{75.56\%}&  \\
			\bottomrule
		\end{tabular}
	\end{center}
	\caption{The top-1 accuracy on the test set of {\it mini}PPlankton.}
	\label{tplank}
\end{table}
For this dataset, we conduct 5-way experiments with both 1-shot and 5-shot trials on the  $\mathcal{D}_{novel}$ and we use the ResNet18 with cosine-classifier as the baseline (the same as Imprint). We randomly selected 10 classes as the basic training classes, and the remaining classes as the novel classes to evaluate few-shot tasks. As shown in table~\ref{tplank}, we can see that the proposed model with CN\_loss outperforms the baselines by a significant margin, from 72.99\% to 74.84\% in the 5-shot trial. However, to our surprise, the high-order module does not work for this dataset, and even leads to decline of test accuracy. The reason is that phytoplankton images are not "closed-shape" (target and background are separate) like normal images. For example, as shown in Fig.~\ref{plankton},  the object of Biddulphia is interspersed with the background.

\begin{figure}[h]
	\begin{center}
		\subfloat[\quad Imprint]{
			\begin{minipage}[t]{0.24\textwidth}
				\centering
				\includegraphics[width=1.0\linewidth]{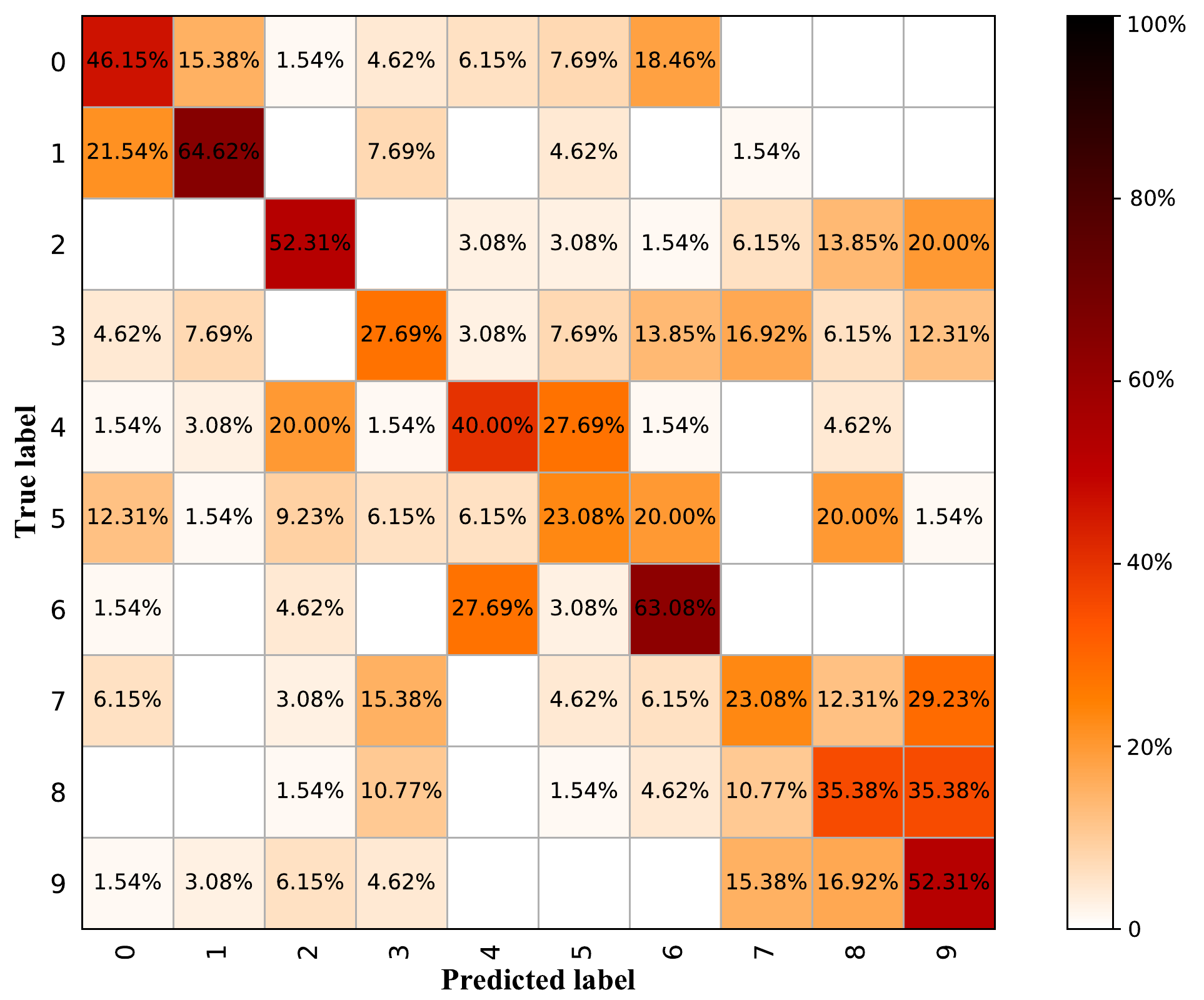} \\	
			\end{minipage}
		}
		\subfloat[\quad Ours]{
			\begin{minipage}[t]{0.24\textwidth}
				\centering
				\includegraphics[width=1.0\linewidth]{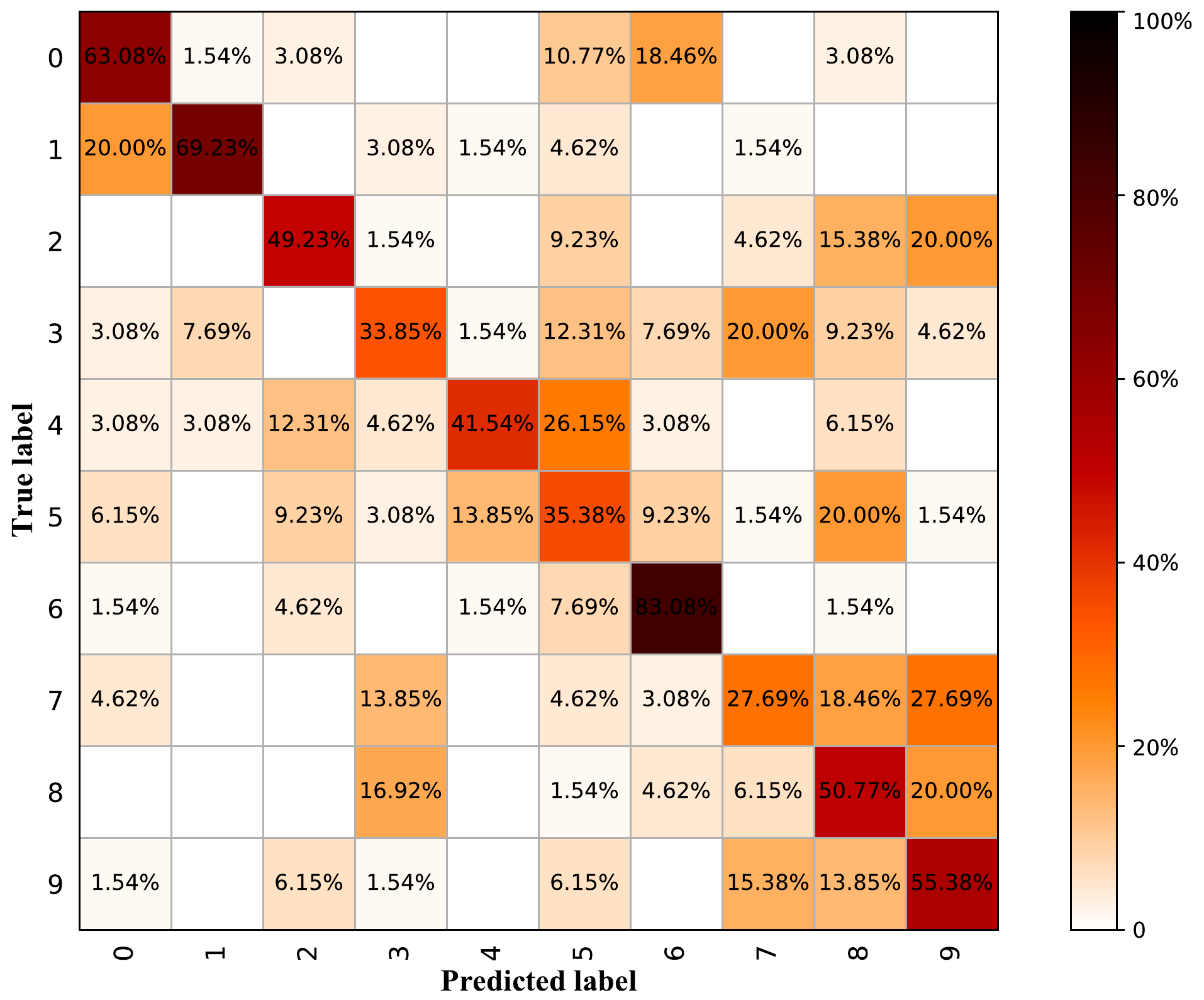} \\	
			\end{minipage}
		}
	\end{center}
	\caption{The confusion matrix of the baseline (ResNet with cosine classifier, also equivalent to Imprint \cite{Brown}) and our methods.}
	\label{confusion}
\end{figure}

\begin{figure}[h]
	\begin{center}
		\includegraphics[width=0.71\linewidth]{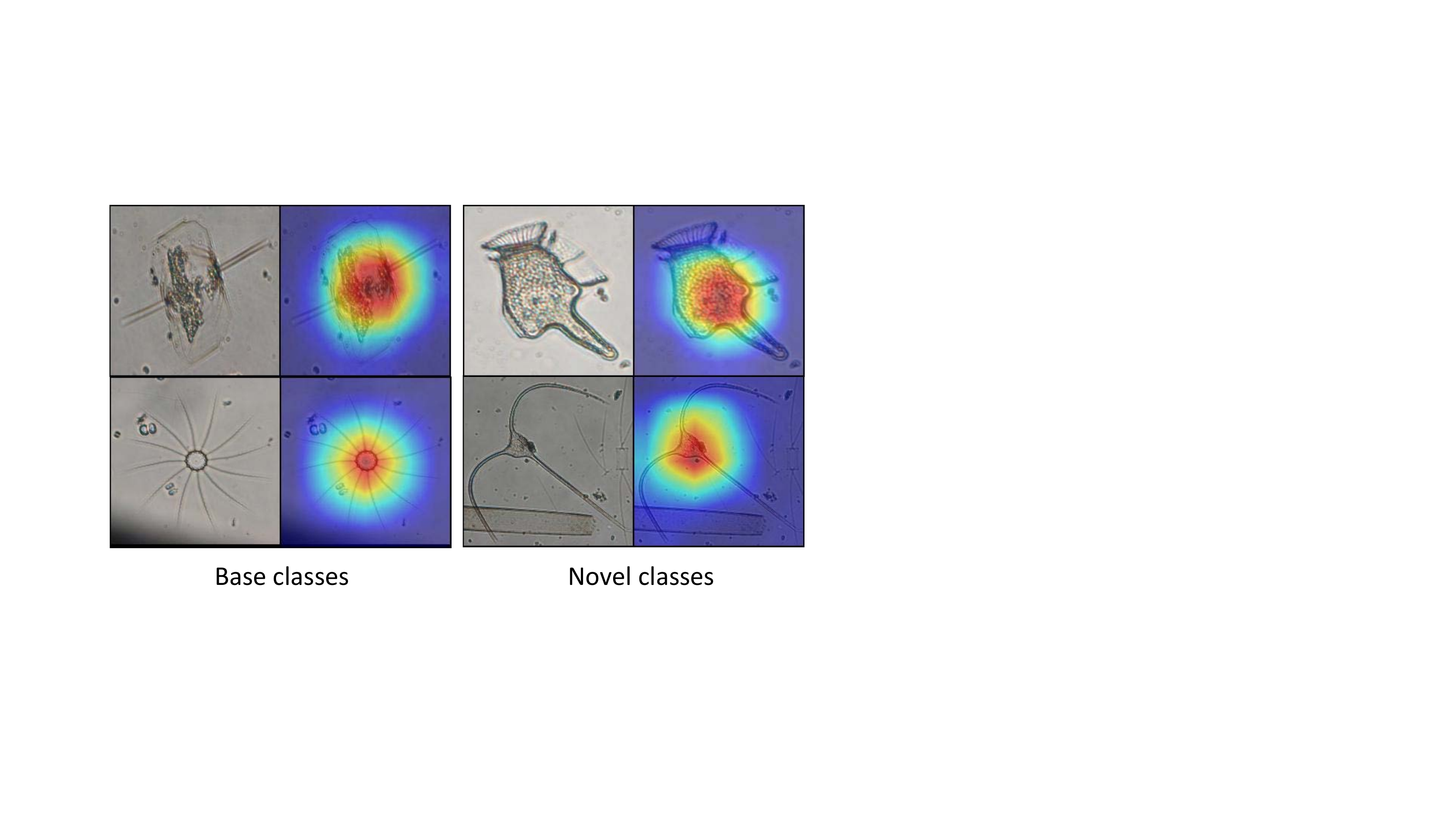}
	\end{center}
	\caption{The focus-area on some examples of the phytoplankton dataset.}
	\label{camplank}
\end{figure}
We further illustrate the improvement of classification performance for each category. Fig.~\ref{confusion} shows the confusion matrices of the baseline and our method on $\mathcal{D}_{novel}$ of {\it{mini}}PPlankton. We can see that our model greatly improves the accuracy of category 1 (pleurosigma-pelagicum) and category 6 (nitzschioides). At the same time, we reduce the possibility of misclassification of category 5 into category 6. However, it is still very challenge for some categories. Moreover, we visualize the focus area of some examples in Fig. ~\ref{camplank}. We can see that our method can capture the key area of the object. It helps the model to extract discriminative features for classification. Fig.~\ref{compareerror} shows the most difficult category pairs. For instance, samples of category 8 are usually classified into category 9. It can be seen from Fig.~\ref{compareerror} that the difference between these categories are very small. Such similarity even confuses marine biologists to distinguish them from each other.

\begin{figure}[h]
	\begin{center}
		\includegraphics[width=0.7\linewidth]{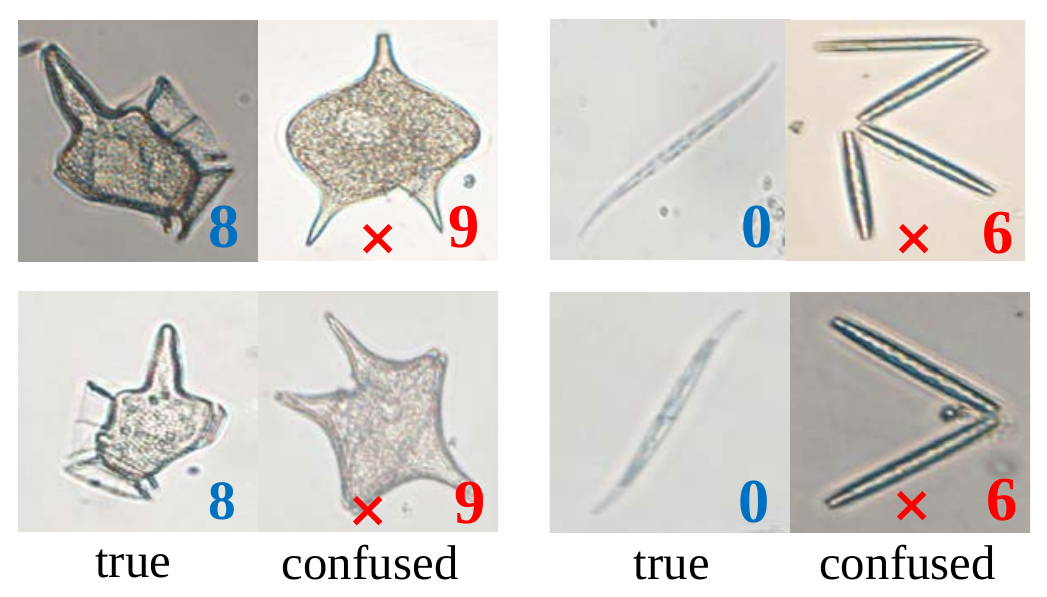}
	\end{center}
	\caption{The blue number in the lower right corner of the image represents the ground-truth category, and the red number represents the wrongly predicted category.}
	\label{compareerror}
\end{figure}

\section{Conclusion}
In this paper, we focus on the challenge of the domain-specific few-shot fine-grained classification problem via exploring the attention features from  a few labeled examples. The Feature Fusion Model and CN\_Loss are our two contributions on mining features for such a challenge task. The fusion model utilizes the focus-area location and high-order integration to generate features from discriminative regions. High-order integration has the ability to capture the intra-parts discriminative information. And Grad-CAM can generate focus-area locations for the novel labeled samples. For few-shot learning, we want to learn a more robust feature extractor through basic training classes. As the fine-grained visual categories are quite similar to each other, we design CN\_Loss to penalize the special samples which are difficult to approach class centers in each iteration.  Furthermore, we build a typical fine-grained and few-shot learning dataset {\it{\textbf{mini}}}{\textbf{PPlankton}} from the real-world application in the area of marine ecological environment. We not only build a few-shot phytoplankton dataset but also design an universal model to accomplish the few-shot classification task of natural images and phytoplankton images in the real-world industrial applications. Extensive experiments are carried out to investigate the effects of these proposed modules. We believe that our method is a valuable complement to few-shot classification problem and the new {\it{\textbf{mini}}}{\textbf{PPlankton}} is attractive for the marine industrial applications.

	\section*{Acknowledgment}
The authors would like to thank anonymous referees for their useful comments and editors for their work. The authors gratefully thank the GPU computation support from Center for High Performance Computing and System Simulation, Pilot National Laboratory for Marine Science and Technology (Qingdao).


\bibliographystyle{Bibliography/IEEEtranTIE}
\bibliography{Bibliography/IEEEabrv,Bibliography/xu}\ 
\end{document}